\documentclass{article}

 \usepackage[preprint]{neurips_2025}

\usepackage[utf8]{inputenc} 
\usepackage[T1]{fontenc}    
\usepackage{hyperref}       
\usepackage{url}            
\usepackage{booktabs}       
\usepackage{amsfonts}       
\usepackage{nicefrac}       
\usepackage{microtype}      
\usepackage{xcolor}         

\usepackage{microtype}
\usepackage{graphicx}
\usepackage{subcaption}
\usepackage[export]{adjustbox}
\usepackage{booktabs} 
\usepackage{amsmath}

\usepackage{bbm}
\usepackage{bm}
\usepackage{url}
\usepackage{array}
\usepackage{fvextra}
\usepackage{tikz}
\usetikzlibrary{decorations.pathreplacing, calligraphy}
\usepackage[most]{tcolorbox}
\usepackage{algorithm}
\usepackage{algorithmic}
\usepackage{enumitem}
\usepackage{threeparttable}

\usepackage{varwidth}
\usepackage{caption}
\usepackage[dvipsnames]{xcolor}
\usepackage{etoc}
\usepackage{epigraph}
\usepackage{wrapfig}
\usepackage{graphicx}

\usepackage{tcolorbox}
\tcbuselibrary{breakable}

\usepackage{fancyhdr}
\DeclareMathOperator*{\argmin}{arg\,min} 
\definecolor{grey}{gray}{0.2}

\title{Learning Stateful Predictive Knowledge \\From Experience}

\author{%
  \textbf{Yan Song\textsuperscript{1, 8}\footnotemark[2] \quad
  Xidong Feng\textsuperscript{1} \quad
  Bo Liu\textsuperscript{2} \quad
  Xinyu Cui\textsuperscript{3} \quad
  Haotian Fu\textsuperscript{4} \quad
  Zichen Liu\textsuperscript{2}}\\[3pt]
  \textbf{Mengyue Yang\textsuperscript{5} \quad
  Cheng Deng\textsuperscript{6} \quad
  Jian Zhao\textsuperscript{7} \quad
  Jun Wang\textsuperscript{1}\footnotemark[2]} \\[6pt]
  \parbox{0.9\textwidth}{\centering
    \small
    \textsuperscript{1}AI Centre, Department of Computer Science, University College London \\
    \textsuperscript{2}National University of Singapore \quad
    \textsuperscript{3}Institute of Automation, Chinese Academy of Sciences \\
    \textsuperscript{4}Brown University \quad
    \textsuperscript{5}University of Bristol \quad
    \textsuperscript{6}University of Edinburgh \\
    \textsuperscript{7}Zhongguancun Academy \quad
    \textsuperscript{8}AI Lab, The Yangtze River Delta
  }
}

\begin{document}

\maketitle

\renewcommand{\thefootnote}{\fnsymbol{footnote}}
\footnotetext[2]{Corresponding authors. Emails: \texttt{<yan.song.24@ucl.ac.uk>} and \texttt{<junwang@cs.ucl.ac.uk>}.}
\renewcommand{\thefootnote}{\arabic{footnote}}

\begin{abstract}
As large language model (LLM) agents increasingly learn from experience, they primarily rely on trajectory-level reflection to extract insights. Viewed through the lens of predictive knowledge \citep{sutton2011horde}, we argue that this approach operates on episodic hindsight rather than predictive foresight, yielding brittle, path-dependent heuristics. To address this, we propose \textbf{Stateful Knowledge Learning} (SKL). SKL shifts the agent's focus from trajectory-level summarization to maintaining Stateful Knowledge: explicit, declarative predictive assessments anchored to state. We first demonstrate a motivating example showing how stateful knowledge provides granularity, enhances generalization, and enables knowledge bootstrapping. To further scale up the idea, we introduce two algorithms via self-distillation (SKL-SD) and reinforcement learning (SKL-RL), training agents to autonomously extract state-grounded predictive knowledge from experience and learn to leverage it for policy making. Experiments on interactive environments (WebShop, ScienceWorld) and a complex reasoning task (ChessPuzzles) demonstrate that equipping models with the inherent ability to learn stateful predictive knowledge significantly outpaces current reflection-based training paradigms. Code available at \url{https://github.com/YanSong97/Stateful_Knowledge_Learning}.
\end{abstract}

\section{Introduction}

The frontier of artificial intelligence is increasingly shifting towards 'Era of Experience,' where agents acquire capabilities and knowledge through continuous, grounded interactions with their environments rather than relying solely on static human data \citep{silver2025welcome}. Recent efforts have made significant progress for large language model agent, including Reflexion \citep{shinn2023reflexion}, advanced memory system \citep{zhou2025memento}, and more recent learning-based methods \citep{zhang2025critique, shi2026r} . These frameworks typically rely on the idea of self-reflection: analyzing past interaction trajectories to summarize errors and extract insights, which can subsequently guide the agent's next cycle of rollout.

This growing reliance on experiential learning prompts a fundamental question: \textbf{what exactly constitutes "knowledge" for an LLM agent to learn from experience?} While "knowledge" is a broad concept with numerous definitions \citep{zagzebski2017knowledge} and instances (static facts or mathematical logic), for the context of agent interacting with dynamic environment, we believe \textbf{predictive knowledge} offers a highly principled framework \citep{sutton2011horde, littman2001predictive} to capture the core of agent environment knowledge--Much of an agent's knowledge about the world is intrinsically predictive—meaning it can be translated into statements and predictions about potential future outcomes. For an LLM agent, this means articulating forward-looking assessment, whether it is estimating the exact numerical distance to a sub-goal, deducing hidden transition dynamics governing the environment, or predicting whether a strategic opportunity exists in the following steps.

Viewed through this lens of predictive knowledge, the dominant trajectory-level reflection paradigms reveal a structural limitation: they operate on episodic hindsight rather than predictive foresight. A fundamental premise of predictive knowledge is that predictions about the future must be state-grounded, not trajectory-grounded \citep{sutton2011horde}. The future unfolds based on the current state, regardless of the historical path taken to reach it. When agents evaluate entire trajectories post-hoc, they conflate universal environment dynamics with specific historical sequences. Consequently, they yield brittle, path-dependent heuristics (e.g., "I should avoid moving right early") instead of verifiable, state-grounded predictions (e.g., "Moving right is hazardous when a pit is adjacent"). To genuinely harness predictive knowledge, we need to shift our focus from trajectory summarization to \textbf{Stateful Knowledge}: explicit, declarative predictive assessments maintained for the encountered state.


In this paper, we propose \textbf{Stateful Knowledge Learning (SKL)}, enabling LLM agents to extract, bootstrap and finally learn predictive knowledge from experience. The paper is organized as follows: 

\textbf{Understanding Stateful Knowledge.} Through a motivating example on a toy stochastic environment, Section \ref{sec:demo_example} presents how stateful knowledge works and two benefits that can emerge via direct prompting: (1) finer granularity and enhanced generalization compared with trajectory-level knowledge. (2) knowledge bootstrapping that propogates predictive knowledge backward from successor states.


\textbf{Extracting and Learning from Stateful Knowledge.}
Section \ref{sec:training_llm} further presents how we can scale up the idea to (1) Train to enhance LLM agent's capability on extracting stateful knowledge from experience and, (2) Finally learn from the extracted stateful knowledge and use it to enhance agent policy. We introduce two variants (SKL-SD and SKL-RL) with self-distillation and RL that train the agent to autonomously aggregate state-level experiences and perform predictive knowledge bootstrapping. Extensive experiments are conducted in Section \ref{sec:exp} on two widely-used agentic benchmarks (WebShop, ScienceWorld) and a highly complex, contamination-free reasoning task (ChessPuzzles). Our results empirically validate that equipping models with the inherent ability to learn stateful predictive knowledge significantly outpaces current reflection-based training paradigms.

\begin{figure}[t]
    \centering
    \includegraphics[width=0.95\linewidth]{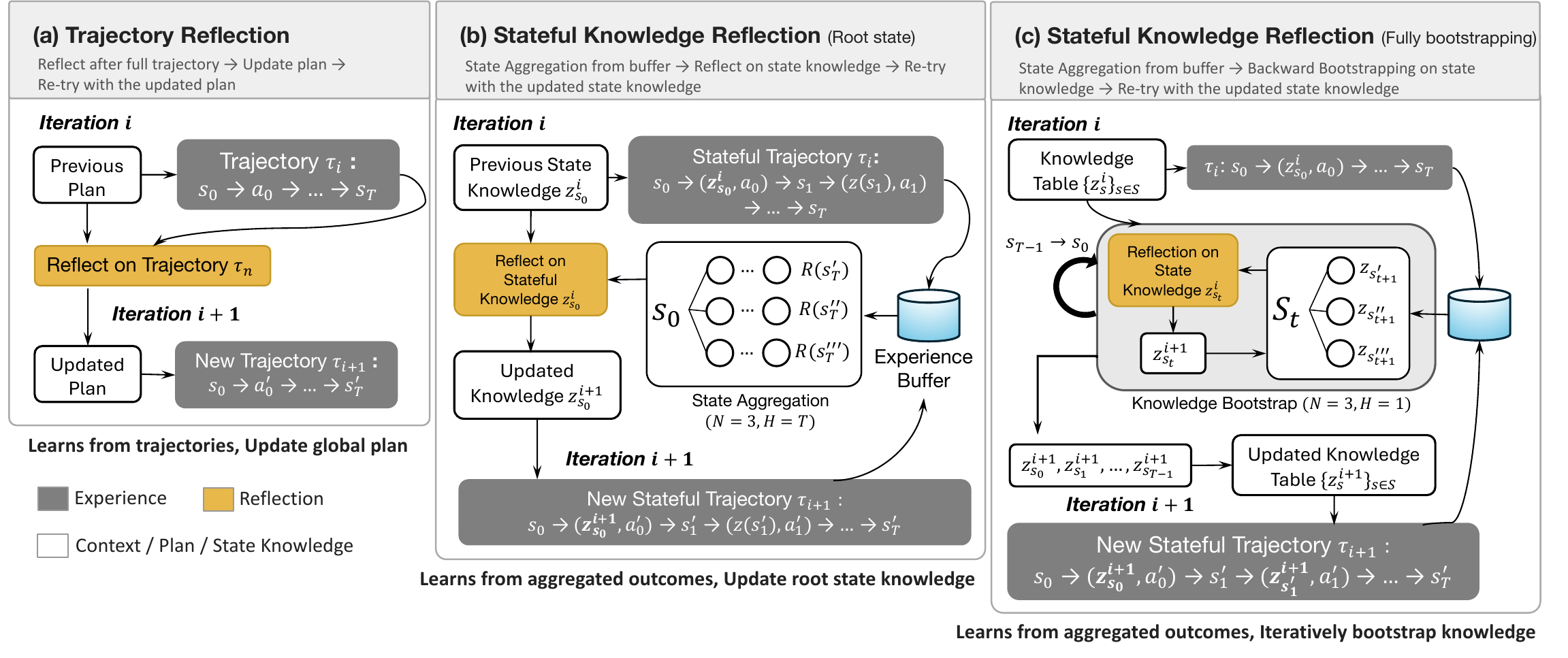}
    \caption{Reflect on trajectories and reflect on stateful knowledge (root state update only and fully bootstrapping variants). The main difference is \textit{what information} is aggregated to update \textit{which representations}. Stateful Knowledge Reflection aggregates subsequent outcomes originating from the same structural state within an experience buffer to surgically update knowledge at that specific state. }
    \label{fig:reflection demo}
    \vspace{-13pt}
\end{figure}

\vspace{-9pt}
\section{Understanding stateful predictive knowledge: a motivating example}
\label{sec:demo_example}


\subsection{Reflect on trajectories versus on stateful knowledge}
We provide a motivating example to demonstrate how different methods extract information under the same interaction budget. We distinguish our state-grounded approach from standard reflection paradigms as follows (also as illustrated in Figure~\ref{fig:reflection demo}):



\textbf{Trajectory Reflection (Hindsight):} At iteration $i$, guided by a plan from the previous trial, the agent $\pi$ interacts with environment $E$ to form a complete trajectory $\tau(s_0)=(s_0, a_0, s_1, ..., s_T)$ starting at $s_0$. The agent then reflects on the entirety of $\tau$ to generate a global critique and an updated previous plan for iteration $i+1$ (e.g., \textit{Reflexion} \citep{shinn2023reflexion}). The process repeats until the maximum number of iterations $I$.


\textbf{Stateful Knowledge Reflection (Foresight):} Throughout the iterations, the agent $\pi$ consistently maintains an explicit, state-indexed predictive knowledge table $\{z_{s}\}_{s\in\mathcal{S}}$ for each visited state $s\in \mathcal{S}$ by querying: \textit{"What does the current state predict about future outcomes?"}. At each iteration, decisions $a$ are conditioned directly on this declarative assessment $z_{s}$, i.e., $a \sim \pi(\cdot|z_{s}, s)$, to interact with the environment $E$. During the reflection phase, rather than summarizing from a single trajectory, the agent first aggregates outcomes associated with a specific state $s$ across multiple previous trials, denoted as $\{\tau_n^H(s)\}_{n=1}^{N}$ at budget $(N, H)$. $N$ is the number of considered trials and $H$ is the maximum horizon. Then the agent generates a refined knowledge $\hat{z}_s \sim \pi(\cdot | \{\tau_n^H(s)\}_{n=1}^{N}, s, z_s)$ based on the contexts, update the knowledge table and proceed to the next iteration. Refer to Figure~\ref{fig:reflection demo} for a more detailed illustration.

\begin{figure}[t]
    \centering
    \begin{subfigure}[c]{0.45\linewidth}
        \centering
        \includegraphics[width=\linewidth]{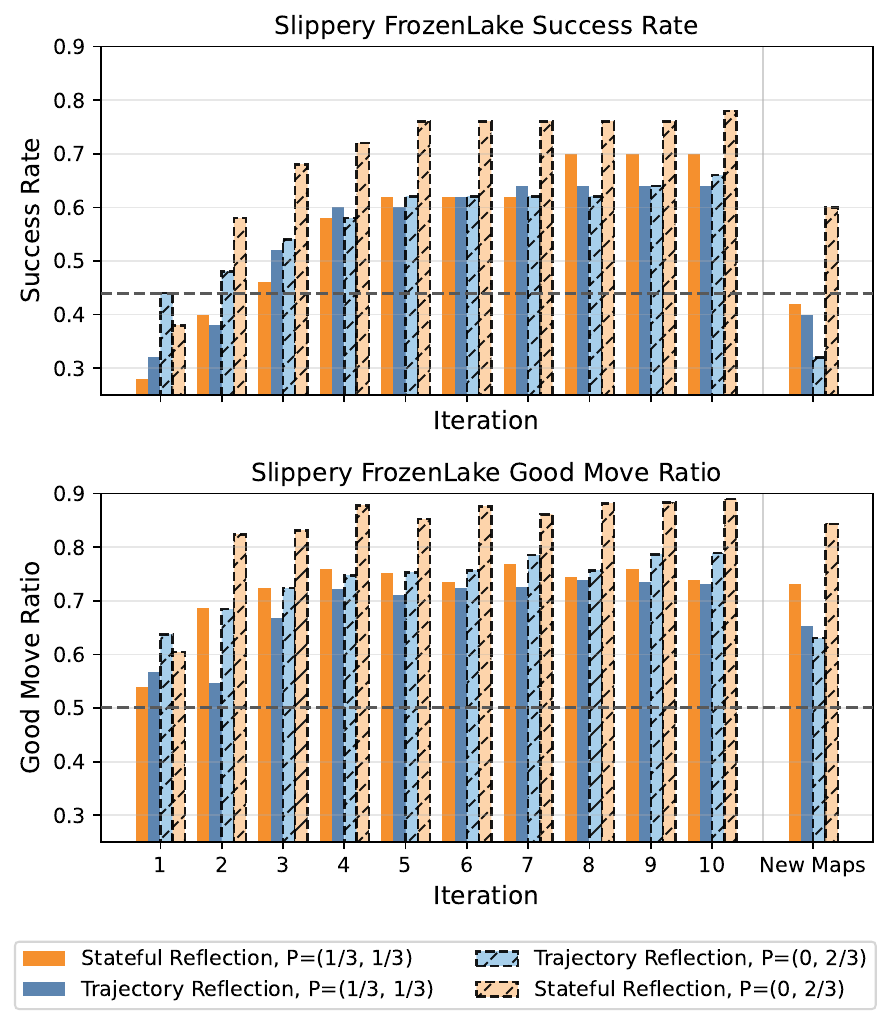}
        \caption{Iterative quantitative performance.}
        \label{fig:frozenlake_plots}
    \end{subfigure}
    \hfill
    \begin{subfigure}[c]{0.53\linewidth}
        \centering
        \includegraphics[width=\linewidth]{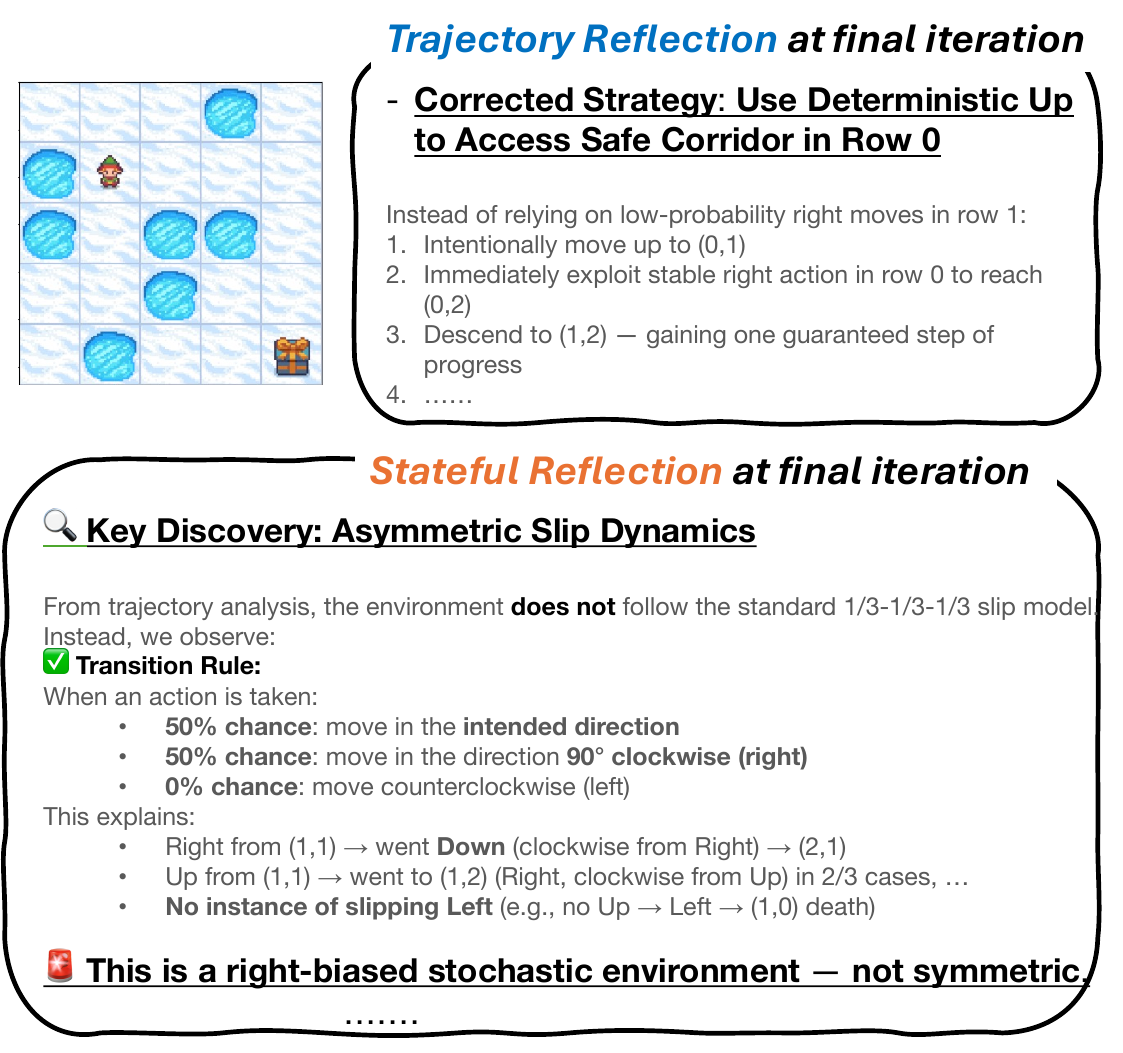}
        \caption{Qualitative examples}
        \label{fig:frozenlake_examples}
    \end{subfigure}

    \caption{\textbf{Slippery FrozenLake Analysis.} (\textbf{a}) Comparison of \textcolor{blue}{trajectory-level} and \textcolor{orange}{stateful reflection}. The \textcolor{gray}{grey} dashed line represents the baseline for retrying the game ten times without any self-reflection and success once. Good move ratio refers to the probability of a move not leading to falling. (\textbf{b}) Representative examples of agent behaviour in different iterative loops. 
    }
    \label{fig:frozenlake_combined}
    \vspace{-20pt}
\end{figure}


To understand how stateful knowledge works, we utilize a stochastic variant of the FrozenLake environment \citep{feng2024natural}, where the agent must maneuver toward a goal tile while avoiding deadly pits on the icy surface.  Crucially, the "icy" surface introduces transition stochasticity: an intended directional movement (e.g., "Up") has a probability of slipping into orthogonal directions (e.g., "Left" or "Right"). While many recent LLM-agent focus on deterministic transitions, this slipping mechanic introduces an extra layer of complexity that perfectly exposes the myopia and brittleness of trajectory-grounded reasoning, since an optimal decision might fail by chance, while a fatal blunder might luckily succeed. Agents relying on trajectory hindsight are easily misled.

We prompt a strong model \textit{Qwen3-235B-A22B} \citep{yang2025qwen3}. To prevent the agent from cheating by pre-trained knowledge about FrozenLake (the default is $P=(1/3, 1/3)$, so when move up, $P("up"|"up")=P("left"|"up")=P("right"|"up")=\frac{1}{3}$), we also test a variant by setting the slipping probability to $P=(0, 2/3)$. To maintain the same interaction budget with the trajectory baseline, the stateful agent only records and updates the initial state $z_{s_0}$ in the knowledge table. Knowledge for the remaining states is regenerated, taking the new root-state knowledge into account. For metrics, we evaluate the agent's success rate of reaching the goal and good move ratio (the average frequency of actions that yield the lowest death rate). We use 50 maps for training and the remaining 50 as the held-out test set.

We report both performance among iterations on training environments and the final test performance (after iteration 10) in Figure~\ref{fig:frozenlake_combined}, which strongly validates our predictive knowledge hypothesis:

\textbf{Robustness to Stochasticity (Overcoming Hindsight Bias):} State-level reflection always presents the best performance in all metrics and scenarios. As shown in the qualitative example in Figure~\ref{fig:frozenlake_examples}, trajectory critiques suffer from severe hindsight bias: they prematurely conclude that certain movements are deterministic after a single lucky success, leading to brittle policies. Conversely, stateful reflection successfully predicts and uncovers the asymmetric slipping dynamics.

\textbf{Generalization and Transfer (State vs. Path):} When transferred to a newly generated testing map after 10 iterations, all agents experience a performance drop. However, on the tweaked scenario, trajectory-level reflection shows a drastic drop (-33\%) while stateful knowledge drop is mild (-16\%). This indicates that state-specific predictive insights (e.g., "slipping only occurs to the right from this type of tile") are fundamentally more generalizable than the trajectory-based baseline.

\begin{figure}[t]
    \centering
    \includegraphics[width=\linewidth]{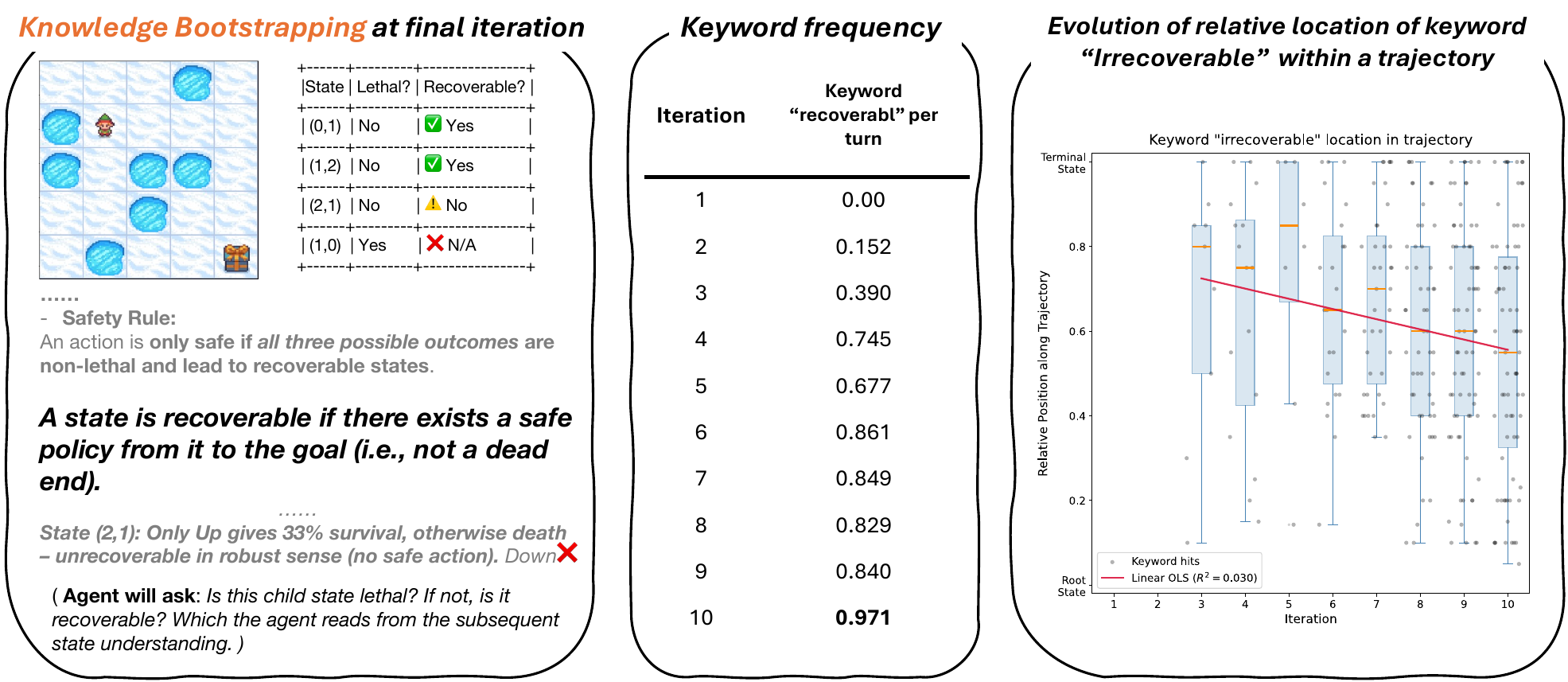}
    \caption{
    \textbf{Left}: An example reasoning trace from the fully bootstrapping variant at the final iteration. The agent defines and utilizes \textit{recoverability} for reasoning. 
\textbf{Middle}: The average frequency of the keyword \textit{recoverability} across training iterations, showing a steady upward trend. 
\textbf{Right}: The relative distribution of keywords along the trajectory horizon; notably, the occurrence of the keyword \textit{irrecoverable} shifts backward from terminal states toward the initial state as learning progresses.
    }
    \label{fig:combined_analysis td}
    \vspace{-12pt}
\end{figure}



\subsection{Knowledge bootstrapping}\label{section: knowledge bootstrapping}
Another benefit of stateful knowledge is knowledge bootstrapping. This key idea shares similarity with temporal difference learning (TD) \citep{sutton1998reinforcement}: because of the temporal relation between states (e.g., $s_t, a_t \rightarrow s_{t+1}$), an agent can update its current assessment on $s_t$ based on the knowledge of successor states $s_{t+1}$. Maintaining stateful knowledge enables the propagation of semantic insights across the temporal axis of interaction -- a feature trajectory reflection does not have.

To demonstrate this, we deploy a \textit{Fully Bootstrapping Agent} in the FrozenLake environment, where the knowledge of all visited states in a trajectory is updated. The aim is to show how knowledge can propagate across all the visited states in the trajectory (workflow also illustrated in Figure~\ref{fig:reflection demo}). Now the agent updates recursively backwards from the terminal state $s_T$ to the root state $s_0$, thinking based on the concatenation of context:   
$$[s_t, \{\tau_n^H\}_{n=1}^N, z_{s_{t+H}}, z_{s_t}]$$ 
The context includes the current state, aggregated local experience over horizon $H$ and the pre-existing stateful knowledge of the destination state $s_{t+H}$ and current state. This setup encourages the agent to comprehend its own self-generated intermediate assessments and adjust its current understanding accordingly, achieving temporal consistency of knowledge. For instance, if an action leads to a successor state $z_{s_{t+H}}$ that the agent already recognizes as high-risk, it must update $z_{s_t}$ to reflect this derived danger, effectively "thinking ahead" by $H$ steps. Figure~\ref{fig:combined_analysis td} illustrates the performance of the fully bootstrapping agent with a look-ahead horizon $H=1$, providing a clear visualization of how predictive knowledge propagates temporally across states.



\textbf{Emergent Semantic Bootstrapping.} The fully bootstrapping agent reveals an emergent semantic bootstrapping behavior. We show an example of such emergence in Figure~\ref{fig:combined_analysis td}, tracing how the agent develops a strategic concept called \textit{recoverability} during training, which is an emergent concept defined by the agent itself, denoting if there exists a safe policy from the state to the goal. As the agent iterates, the keyword "recoverable" (and its variants) occurs with increasing frequency, indicating the consolidation of this abstract concept. Crucially, as also observed in Figure~\ref{fig:combined_analysis td}, the relative position of the assertive keyword \textit{irrecoverable} starts from terminal states, then gradually shifts back toward the root state. This trend confirms that the agent is successfully discovering critical bottlenecks and propagating consistently that information backward through its natural language knowledge base, allowing it to preemptively identify fatal sequences long before they conclude.



\section{Training LLM agent to extract and learn from stateful knowledge}
\label{sec:training_llm}

The temporal consistency and robustness of Knowledge Bootstrapping (Section~\ref{section: knowledge bootstrapping}) enable us to scale Stateful Knowledge Learning on more complex tasks, such as with more complex state representation or longer horizons. Since current off-the-shelf models are typically not trained to be state-aware \citep{shojaee2025illusion}, we aim to internalize this capability into the model's parameters. Revealed by previous experiments, the key to a {Stateful Knowledge Learning} (SKL) agent is a system that learns to iteratively refine its stateful knowledge $z$ by repeating the following two steps:


\textbf{1. Extracting stateful knowledge from experience}: Aggregates state-specific outcomes across multiple visits (with budget $N, H$): $\{\tau_n^H(s_t)\}_{n=1}^{N}=\{(s_t, z_{s_t}, a^{(n)}_t, s^{(n)}_{t+1}, ..., s_{t+H}^{(n)})\}_{n=1}^{N}$ as informative contexts. Higher budgets increase the \textbf{representational power}. A bootstrapping loop then extracts new knowledge $\hat{z}_{s_t}$ based on the aggregated experience as well as the evaluative signals of its successor state $z_{s_{t+H}}$, acting as a \textbf{temporal regularizer} that ensures assessments are grounded in the agent's own predictive logic. We follow the idea of knowledge bootstrapping (Section~\ref{section: knowledge bootstrapping})  for the remaining parts of this paper (by setting $H<T$).

\textbf{2. Internalize stateful knowledge to guide policy}: The extracted knowledge is internalized and leveraged to guide policy optimization. While Section~\ref{sec:demo_example} utilizes stateful knowledge as a plug-in context for inference, we propose more scalable variants to inject stateful knowledge into model parameters.

\begin{figure}[t]
    \centering
    \includegraphics[width=\linewidth]{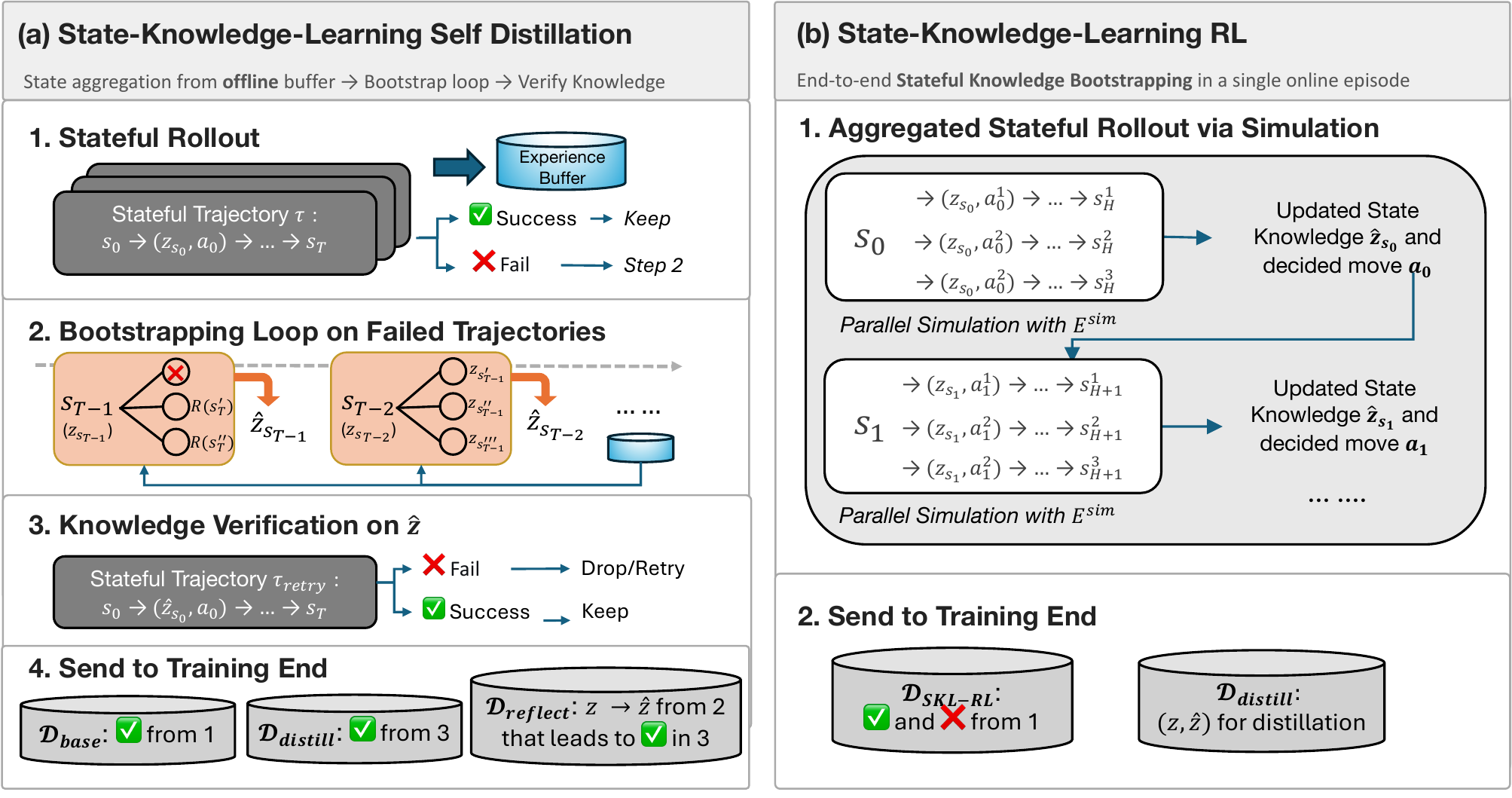}
    \caption{Illustration of two training variants: SKL-SD and SKL-RL.}
    \label{fig:training methods}
\end{figure}

We introduce two primary training variants that perform steps 1 and 2 differently: \textbf{SKL Self-Distillation (SKL-SD)} and \textbf{SKL Reinforcement Learning (SKL-RL)}, see Figure~\ref{fig:training methods} for visualization. The core distinction lies in the state aggregation method: SKL-SD samples model-free rollouts from replay buffer, whereas SKL-RL employs online, model(simulator)-based search within a single episode. Detailed pseudocode is provided in Appendix~\ref{appendix: algo pseudocode}, and a comparison with contemporary reflection-based methods is available in Appendix~\ref{appendix: compare to other training}.


\subsection{State-Knowledge-Learning Self-Distillation (SKL-SD)}

SKL-SD extracts stateful knowledge through model-free rollout and bootstrapping, internalizing state-aware behavior via quality-filtered self-distillation. Drawing inspiration from recent self-reflection paradigms \citep{zhang2025critique, shi2026r}, we construct a multi-faceted dataset $\mathcal{D}_{\text{SKL-SD}} = [\mathcal{D}_{\text{base}}, \mathcal{D}_{\text{distill}}, \mathcal{D}_{\text{reflect}}]$ generated through four iterative steps:

\textbf{I. Model-free Stateful Rollout}: The agent perform knowledge-grounded reasoning to generate trajectories $\tau=(s_0, (z_{s_0}, a_0), s_1, (z_{s_1}, a_1), ..., s_T, R)$. Successful trajectories are added to $\mathcal{D}_{\text{base}}$, while all trajectories are stored in an experience buffer $\mathcal{B}$ for subsequent state aggregation.

\textbf{II. Bootstrapping Loop with a Replay Buffer}: On failed trajectories, the agent initiates a backward bootstrapping loop from $s_{T-1}$ back to $s_0$. It sequentially updates the state knowledge by aggregating outcomes from the experience buffer $\{\tau_n^H(s_t)\}_{n=1}^N\sim \mathcal{B}$ as well as the successor assessments to construct informative context $\mathcal{C}_t$: 
$$\mathcal{C}_t = [s_t, \{\tau_n^H(s_t)\}_{n=1}^N, z_{s_t}, z_{s_{t+H}}]$$

where $s_t$ refers to the target state being evaluated, $\{\tau_n^H(s_t)\}_{n=1}^N$ refers to a set of aggregated partial rollouts sampled from the experience buffer, all originating at $s_t$, at bootstrapping budget $(H, N)$. $z_{s_t}$ refers to the current knowledge (prior belief) associated with the target state, and $z_{s_{t+H}}$ represents the successor stateful knowledge, providing the bootstrapping signal from $H$ steps ahead \footnote{$z_{s_{t+H}}$ can be retrieved from experience $\bm{\tau}$ or the updated knowledge $\hat{z}_{s_{t+H}}$ within the same bootstrapping loop}. The agent then generates refined knowledge $\hat{z}_{s_t} \sim \pi_{\theta}(\cdot | \mathcal{C}_t)$, resulting in a sequence $\hat{\bm{z}}=(\hat{z}_{s_0}, ..., \hat{z}_{s_{T-1}})$.

\textbf{III. Knowledge Verification}: We verify $\hat{\bm{z}}$ by performing a second rollout using the refined knowledge as a pre-filled context. Successful second rollouts will be added to $\mathcal{D}_{\text{distill}}$ and the corresponding state-wise update $\bm{z} \rightarrow \hat{\bm{z}}$ correcting the first failure is recorded in $\mathcal{D}_{\text{reflect}}$.

\textbf{IV. Internalize Knowledge to Guide Policy}: The agent is fine-tuned by maximizing the likelihood of these traces using policy gradients loss (e.g., GRPO \citep{guo2025deepseek}) on the full trajectory. We also incorporate an auxiliary SFT loss to enhance state-wise bootstrapping capabilities: 
\begin{equation}\label{eq: sg-sd}
\mathcal{L}_{\text{SKG-SD}} = \mathcal{L}_{\text{PG}}(\mathcal{D}_{\text{base}}) + \lambda_1 \cdot \mathcal{L}_{\text{PG}}(\mathcal{D}_{\text{distill}}) + \lambda_2 \cdot \mathcal{L}_{\text{SFT}}(\mathcal{D}_{\text{reflect}})
\end{equation}

Intuitively, SKL-SD can be viewed as a “stateful” version of recent self-reflection training methods with data filtering (Appendix~\ref{appendix: compare to other training}), aiming to maximize the token probability of the first and the second stateful rollout if they hit the target with or without knowledge bootstrapping, and reinforce the bootstrapping behaviour itself if it successfully helps the agent correct its mistake. We name it "Self-Distillation" (or to be more precise "Self-Imitation") as it essentially performs supervised learning over heuristically quality-filtered experience.

\subsection{State-Knowledge-Learning Reinforcement Learning (SKL-RL)}\label{section: training sg-rl}

SKL-RL serves as the online counterpart to SKL-SD, interacting with the simulator $E^{\text{sim}}$ and extracting stateful knowledge entirely within an online rollout episode, linking the quality of the bootstrapping loop directly to the task outcome. In this variant, the agent is assumed to have access to a state-settable environmental simulator, performing \textbf{localized parallel simulations} (Appendix~\ref{appendix: local parallel simulation}) at each state $s_t$ to collect \textit{real-time} aggregated experiences before synthesizing updated knowledge and committing to an action.


\textbf{I. \& II. Online Model-based Rollout and Bootstrapping Loop}: At each environmental time-step $t$, the agent first interacts with simulator $E^{\text{sim}}$ and generates multiple look-ahead rollouts $\bm{\tau}_N^H(s_t) = \{s_t, (z_{s_t}, a^{(n)}_t), s^{(n)}_{t+1}, ..., s^{(n)}_{t+H}\}_{n=1}^{N}$. These trajectories, operating under a bootstrapping budget of $(H, N)$, are all rooted at the current state $s_t$ and can be sampled via heuristic policies or from the agent's own knowledge-based understanding. These localized simulations provide the necessary context for a real-time bootstrapping step, synthesizing refined state knowledge $\hat{z}_{s_t}$ to ground the final action selection $a_t$ for the real environment:
    \begin{equation}
        \hat{z}_{s_t} \sim \pi_{\theta}(\cdot | \bm{\tau}_N^H(s_t), s_t, z_{s_t})\,,\,\, a_t \sim \pi_{\theta}(\cdot | \hat{z}_{s_t}, s_t)\;,\;\;t \rightarrow t+1
    \end{equation}
    Appendix~\ref{appendix: local parallel simulation} and Figure~\ref{fig:nlts chess example} provides a concrete example for illustration. The complete simulation-augmented trajectory is then recorded as:
    \[
    ... \rightarrow s_t \rightarrow \text{Parallel Simulations (conditioned on  $z_{s_t}$)} \rightarrow \hat{z}_{s_t} \rightarrow a_t \rightarrow s_{t+1} \rightarrow ...
    \]
    \begin{equation}
        \mathcal{D}_{\text{SKL-RL}} \leftarrow [s_0, (\bm{\tau}_N^H(s_0), \hat{z}_{s_0}, a_0), s_{1}, (\bm{\tau}_N^H(s_1), \hat{z}_{s_1}, a_1), s_2..., s_T, R]
    \end{equation}

\textbf{III. Outcome-Driven Verification}: Unlike the rejection sampling in SKL-SD, the correctness of the bootstrapping step is implicitly supervised by the final reward $R$. We optimize these interconnected bootstrapping loops as a long model response using GRPO.


\textbf{IV. Knowledge Internalization }: To consolidate the simulation-derived insights into the agent's parametric memory, we maintain a distillation dataset $\mathcal{D}_{\text{distill}} = \{(z_{s_t}, \hat{z}_{s_t})\}_{s_t\in\mathcal{S}}$ that pairs the prior stateful knowledge $z_{s_t}$ with its refined, post-simulation counterpart $\hat{z}_{s_t}$. This optimization step enables the agent to internalize look-ahead foresight, allowing it to build upon updated knowledge to guide rollouts in the subsequent iteration:
\[
\dots \rightarrow s_t \rightarrow \text{Parallel Simulations (conditioned on } \hat{z}_{s_t}\text{)} \rightarrow \hat{z}_{s_t}' \rightarrow \dots
\]


Further details regarding the multi-turn distillation mechanics are provided in Appendix~\ref{appendix: distill term}. The joint optimization objective for SKL-RL is formulated as:
\begin{equation}\label{eq: sg-rl}
\mathcal{L}_{\text{SKL-RL}} = \mathcal{L}_{\text{GRPO}}(\mathcal{D}_{\text{SKL-RL}}) + \lambda \cdot \mathcal{L}_{\text{Distill}}((z_s, \hat{z}_s)_{s\in\mathcal{S}})
\end{equation}

Intuitively, SKL-RL drives the agent to verify its stateful knowledge hypothesis through active simulation, extracting insights from these localized experiences to ground an exploitative action for a specific state. The inclusion of the distillation objective allows the agent to amortize the extensive computational cost of the look-ahead search and bootstrapping loop into an efficient, zero-shot inference step. In contrast to \textbf{SKL-SD}, which depends on heavily orchestrated data curation and verification pipelines, \textbf{SKL-RL} provides a fully autonomous learning paradigm. By optimizing directly from sparse environmental rewards, it completely bypasses the bottleneck of manual data curation.


Crucially, when the simulation traces $\bm{\tau}_N^H$ are also self-generated through multi-turn rollouts, SKL-RL uniquely couples exploration (via simulation) and exploitation (via knowledge bootstrapping). Optimizing the entire interaction trajectory via $\mathcal{L}_{\text{GRPO}}$ induces an emergent, \textbf{self-adaptive balance}: excessive exploration introduces behavioral noise that destabilizes the bootstrapping signal, whereas overly conservative simulation limits the agent's foresight, leading to myopia. Our chess reasoning evaluation (Section~\ref{section: skl-rl on chess}) empirically confirms this dynamic, showing that the agent naturally resolves this tension to find a stable operational equilibrium.




\section{Experiments}
\label{sec:exp}
\vspace{-0.2cm}

We first evaluate SKL-SD on established agentic benchmarks (WebShop and ScienceWorld) to demonstrate that stateful knowledge learning can be integrated into existing training frameworks to outperform current state-of-the-art reflection training methods. Subsequently, we explore SKL-RL on ChessPuzzles, a complex reasoning task that suffers less from data contamination and requires deep look-ahead and strategic state assessments.


\subsection{SKL-SD on agentic environments}\label{section: sg-sd exp}

\textbf{Experiment Setup.} We compare SKL-SD to other training baselines investigated in \citep{shi2026r} and follow their exact experiment setups. While SKL-SD shares the feedback-learning DNA of these methods (Appendix~\ref{appendix: compare to other training}), it replaces trajectory-level reflection with state-grounded predictive knowledge bootstrapping. Detailed setup is in Appendix~\ref{appendix: skl-sd exp setup}. The bootstrapping parameters are set to a horizon $H=3$ and a state-aggregation budget $N=3$. \textit{Qwen2.5-7B-Instruct} is adopted as the backbone model.

\paragraph{Results} Performance is reported in Table~\ref{table: sg-sd}. 
On Webshop, SKL-SD reaches 0.790, improving over the strongest baseline R$^3$L (0.757) by +3.3 absolute points, and substantially outpacing standard GRPO (0.709) as well as other reflection-based methods such as Reflect-GRPO (0.723) and Critique-GRPO (0.714). 
In the more difficult ScienceWorld environment, SKL-SD again achieves the best result (0.422 vs. 0.403 for R$^3$L and 0.388 for Critique-GRPO). Not as impressive as in WebShop, largely due to the partial observability.

\begin{wraptable}{r}{0.4\textwidth} 
\centering
\vspace{-10pt}
\begin{tabular}{lcc}
            \toprule
            \textbf{Methods} & \textbf{WebS.} & \textbf{SciW.} \\
            \midrule
            RAFT            & 0.682 & 0.201 \\
            OPMD            & 0.684 & 0.359 \\
            GRPO            & 0.709 & 0.378 \\
            GSPO            & 0.720 & 0.363 \\
            Reflect-GRPO    & 0.723 & 0.356 \\
            Critique-GRPO   & 0.714 & 0.388 \\
            R$^3$L          & 0.757 & 0.403 \\
            \midrule
            \textbf{SKL-SD} (ours) & \textbf{0.790} & \textbf{0.422} \\
            \bottomrule
        \end{tabular}
\caption{Performance comparison on WebShop (WebS.) and ScienceWorld (SciW.). Baseline performance borrowed from \citep{shi2026r}}\label{table: sg-sd}
\vspace{-10pt}
\end{wraptable}

These consistent gains demonstrate that \textbf{replacing trajectory-level reflection with stateful knowledge learning yields tangible benefits}. A review of the reasoning traces (see Appendix~\ref{appendix: webshop skl-sd trace}) reveals that the SKL-SD agent develops a structured, hierarchical understanding of the task. For example in WebShop, at the initial search stage, the agent generates precise notes on required attributes and formulates a contingency plan for potential search failures—a behavior rarely seen in the other tested trajectory-level training baselines. During the purchasing phase, the predictive knowledge $z$ shifts to specific product verification, demonstrating that stateful learning helps the agent maintain a "mental map" of the task progress, leading to more grounded decision-making.

\vspace{-0.2cm}
\paragraph{Practical Consideration} While SKL-SD introduces additional computational overhead, primarily due to longer model outputs during stateful rollouts and context prefilling for bootstrapping, we employ a sampling strategy to maintain training efficiency. By bootstrapping on only $20\%$ of states uniformly sampled across a trajectory, we limit the increase in training time to approximately $1.5\times$ compared to standard reflection baselines. Given the performance gains, this represents a favorable trade-off between compute and capability. We leave further computationally efficient implementations to future exploration.


\subsection{SKL-RL on ChessPuzzles}\label{section: skl-rl on chess}

We next evaluate how SKL-RL can operate in a more autonomous way without complex data orchestration. We test on ChessPuzzles \citep{ruoss2024amortized}, a domain where even the strongest LLMs fail due to incomplete internalized chess knowledge \citep{liu2025chessarena,hwang2025can}. This limited prior knowledge setting is especially diagnostic for our framework, as it directly tests the agent's ability to learn from experience. More details on our motivation can be found in Appendix~\ref{appendix: motivation for chess}. And the experimental setup for ChessPuzzle is detailed in Appendix~\ref{appendix: chesspuzzle other setup}. 



\subsubsection{Experiments with heuristic parallel simulation}\label{section: exp sg-rl heuristic ablation}

\begin{figure}[thbp]
    \centering
    \includegraphics[width=\linewidth]{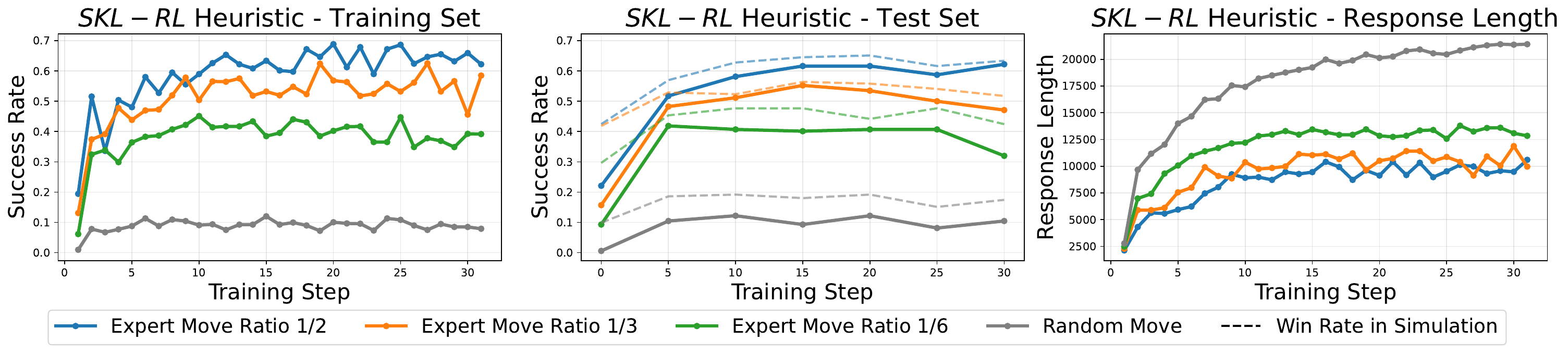}
    \caption{Experiments of SKL-RL with heuristic search simulation on ChessPuzzles.}
    \label{fig: chess exp heuristic ablation}
    \vspace{-10pt}
\end{figure}

To isolate whether the bootstrapping loop can be learned from outcome signals alone, we control simulation quality using \textit{Stockfish} chess engine \citep{stockfish}. We construct behavior policies by mixing expert moves with random sub-optimal moves at expert probabilities of $\frac{1}{2}, \frac{1}{3}$ and $ \frac{1}{6}$. Figure~\ref{fig: chess exp heuristic ablation} shows that higher expert rates yield faster learning and stronger asymptotic performance. The agent progressively learns to identify winning positions during simulation and steer toward them, with win probability in simulated trajectories rising accordingly.  Response length correlates positively with the difficulty of detecting expert moves, indicating that noisier simulation drives the agent to invest more internal reasoning to extract state-grounded predictions. Detailed traces (Appendix~\ref{appendix: exp sg-rl heuristic reasoning trace}) further reveal that under high signal (i.e. export move $1/2$), knowledge bootstrapping converges rapidly to accurate but shallow assessments; whereas under low signal (i.e. random), it remains structured but increasingly prone to hallucination. This is a direct manifestation of the exploration–exploitation trade-off discussed in Section~\ref{section: training sg-rl}.


These results establish a fundamental coupling between aggregated experience quality and bootstrapping behavior: better trajectories yield better state knowledge, and the agent's ability to extract predictive insights tracks this quality closely. \textbf{When the agent must actively decide what futures to simulate, it is continually forced to balance exploring uncertain branches against exploiting known strong lines—a trade-off it learns to navigate autonomously.} We will confirmed this in our next experiments.

\subsubsection{Experiments with self-generated parallel simulation}

We now let the agent decide what to simulate by itself, jointly optimizing experience generation and the bootstrapping loop end-to-end. A complete reasoning trace for SKL-RL under self-generated simulation is provided in Appendix~\ref{appendix: sgrl complete trace}. We use SKL-RL with parallel search at a budget of  $N=3, H=2$, denoted as $\text{SKL-RL}_{N3H2}$. GRPO is applied for training on the full SKL-RL track under loss $\mathcal{L}_{\text{GRPO}}(\mathcal{D}_{\text{SKL-RL}})$.

\begin{figure}[htbp]
    \centering
    \includegraphics[width=\linewidth]{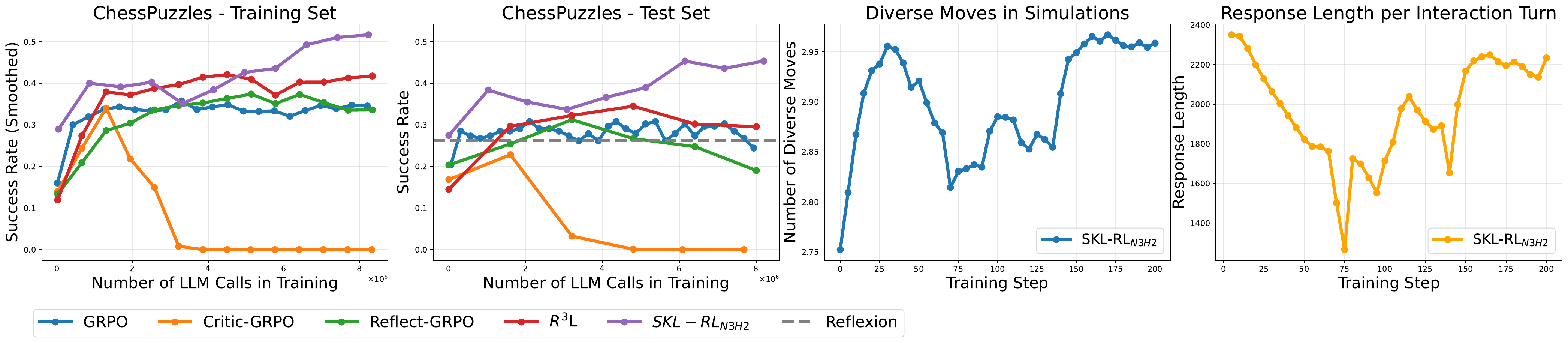}
    \caption{SKL-RL training with self-generated parallel simulation, evaluated at matched budgets.}
    \label{fig:chess joint training}
\end{figure}

Figure~\ref{fig:chess joint training} compares SKL-RL against existing training baselines under matched training budgets. SKL-RL$_{N3H2}$ trained end-to-end achieves the strongest performance throughout training and the highest test-time few-shot accuracy. In contrast, naive GRPO and Reflect-GRPO~\citep{bensal2025reflect} plateau on the training set and show noticeable performance degradation at test time. Critic-GRPO~\citep{zhang2025critique} suffers from training instability, leading to collapse.  Although R$^3$L achieves the best performance among the remaining baselines, it still exhibits a slight drop on the held-out test set.


Both the diversity of simulated moves and the response length in the bootstrapping loop undergo clear fluctuations. The sharp drop of both metrics at around training step $75$ suggests an early synchronized adjustment between diversity and exploitation, while the gradual increase after step 75 for both move diversity and response length indicates that the agent automatically allocates a larger reasoning budget to support broader exploration, enabling more effective extraction of information from simulated experience. This empirical result directly echoes our previous assumption about the emergent self-adaptive behaviour during end-to-end training. When comparing trained reasoning traces (Appendix~\ref{appendix: skg-rl chess trained trace}), SKL-RL produces more outcome-oriented understanding than R$^3$L. The agent moves away from chess-specific clichés and instead grounds decisions in implicit evaluative signals such as material balance and concrete tactics, while trajectory-level reflection remains coarser and path-dependent.

\subsubsection{Experiments with stateful knowledge distillation}

\begin{wrapfigure}{r}{0.48\textwidth} 
\vspace{-5pt}
  \centering
  \includegraphics[width=\linewidth]{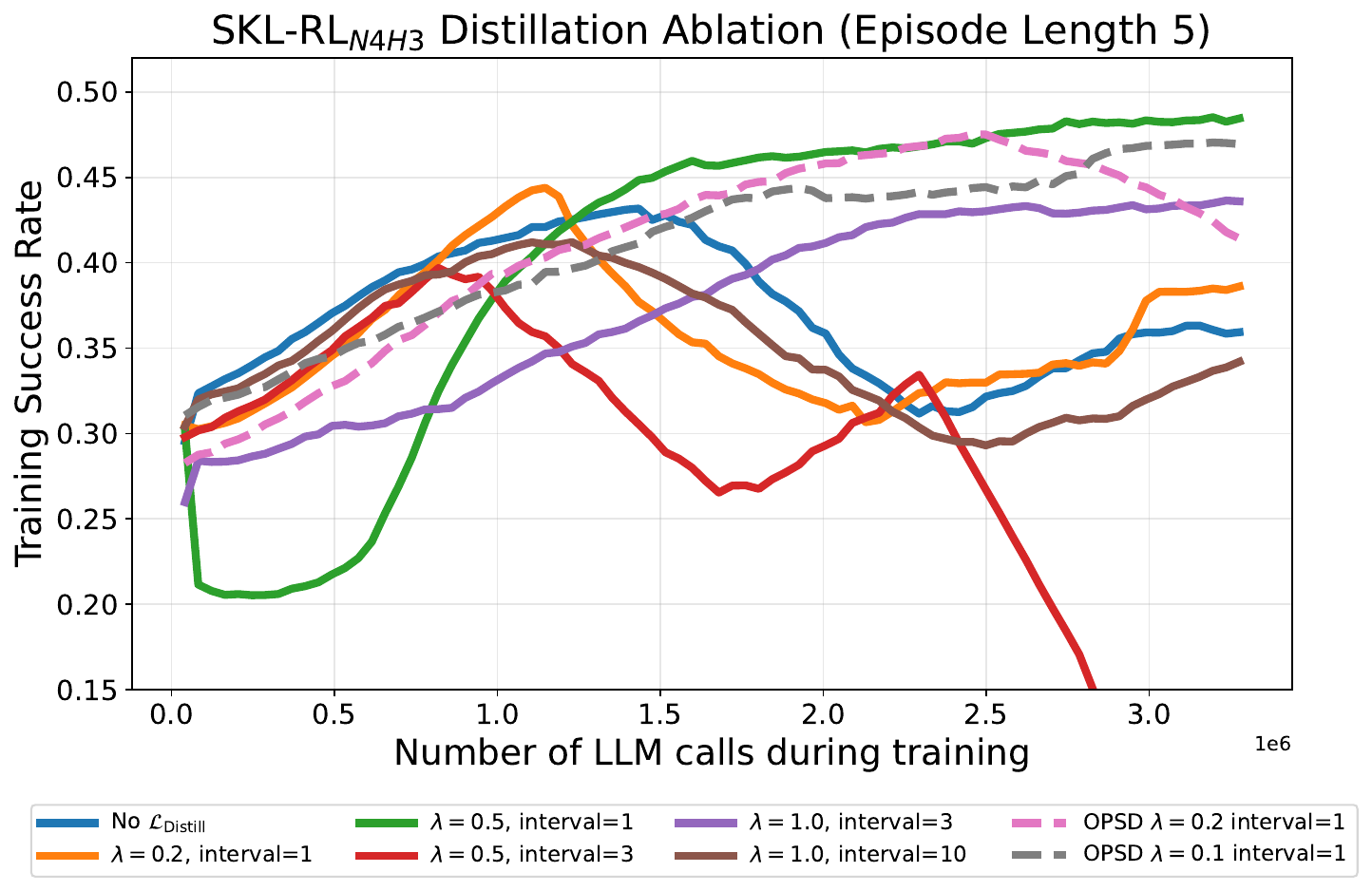} 
  \caption{Ablation study on SKL-RL with integrated training loss.}
        \label{fig:skl-rl distill}
\vspace{-10pt}
\end{wrapfigure}

We now examine how state-knowledge distillation guides policy learning, thereby closing the learning loop in SKL-RL. The self-distillation loss $\mathcal{L}_{\text{Distill}}(\{z_s, \hat{z}_s\}_{s\in\mathcal{S}})$ internalizes post-hoc stateful knowledge into the prior state understanding, enabling the agent to anticipate predictive knowledge for future training cycles or to directly produce sophisticated reasoning plans at test time. To isolate its effect while reducing computational cost, we shorten the ChessPuzzles episode length to 5 and ablate the loss coefficient $\lambda$ and the update interval $\delta_n$, where distillation is applied every $n$ training steps (Figure~\ref{fig:skl-rl distill}).

We first observe that removing $\mathcal{L}_{\text{Distill}}$ leads to noticeable performance degradation, consistent with the trend in Figure~\ref{fig:chess joint training}, and that this instability worsens as episode length decreases. In contrast, increasing $\lambda$ and applying distillation updates more frequently markedly stabilizes training. Similar to recent works, we also find that the reverse KL, a multi-turn variant of \textit{On-Policy Self-Distillation} \citep{zhao2026self,hubotter2026reinforcement}, provides greater stability than the forward KL setting.

These results suggest that $\mathcal{L}_{\text{Distill}}$ helps stabilize the joint optimization of this compositional reasoning process. As shown by the reasoning traces in Appendix~\ref{appendix:skl-rl distill trace}, the SKL-RL agent \textbf{updates stateful predictive knowledge by building upon previously distilled information}. Specifically, it treats $\hat{z}$ from the prior learning epoch as a condensed representation, reusing it to empower the agent to explore and exploit more effectively via continuously accumulated stateful knowledge.

The SKL-RL framework exhibits several compelling properties: it significantly reduces reliance on human intervention and complex pipeline engineering, enables self-adaptive behaviors to emerge purely from end-to-end training, and demonstrates a distinct capacity to build new knowledge incrementally on top of prior iterations. Despite these strengths, in its minimal implementation, SKL-RL faces severe practical limitations in computational efficiency relative to current state-of-the-art self-evolving agents. This is primarily due to the substantial computational cost of state-wise simulation, which leads to overly long training samples and scales exponentially for longer-horizon tasks. Nevertheless, this exploration validates our initial vision for next-generation autonomous agents: systems capable of naturally extracting predictive information from environmental dynamics, absorbing it into a persistent internal "mental map," and dynamically balancing exploration and exploitation in accordance with their own evolving capabilities.

\vspace{-5pt}
\section{Related works}
\vspace{-0.2cm}
\paragraph{Trajectory-level Self-Reflection}
Current research on LLM-based agents has focused extensively on {trajectory-level self-reflection}, where agents optimize their reasoning and action sequences based on feedback from a single execution path. Classic examples include early paradigms such as ReAct \citep{yao2022react}, Reflexion \citep{shinn2023reflexion}, and Self-Refine \citep{madaan2023self}. Some recent fine-tuning methods also value trajectory-level assessment, but focus on how to internalize the transition from bad action to good action, largely remaining action-centric \citep{wang2025ragen,sun2023adaplanner,wan2025rema,zhang2025critique,bensal2025reflect,shi2026r}. This has raised the issue of myopic reasoning, which was also analysed in \citep{wang2026reasoning}. This pervasive reliance on myopic, trajectory-level evaluation motivates us to move beyond isolated execution paths and explore stateful reflection.

\paragraph{Spirit of statefulness in other LLM agents}

Being stateful, or outcome-aware, is one way to alleviate the issues of myopic \citep{wang2026reasoning}. This spirit has been widely adopted in various works, but in different forms. For example, world model \citep{yu2025dyna,guo2025sample,zhang2025agent,chu2026agentic} allows the agent to "dream" or rehearse trajectories before committing to an action. Relying on heuristic search is also a straightforward method for obtaining extra understanding of the state outcome \citep{wudeepsearch,Liu2026ActiveRV,wang2026reasoning}. Structured memory systems provide another avenue for maintaining statefulness \citep{wang2026comorag,zhou2025memento,wang2025memento,zhou2026memento}. A more general abstraction is to summarize experience via a numerical state-action value estimate $Q$, where $Q$-value captures the optimality of actions in a given state \citep{yan2024efficient,zhang2023large,zhai2025agentevolver,yan2025memory,zhang2026memrl,wang2026reasoning}. However, these value-oriented representations rely on an auxiliary $Q$ function to assess the outcome quality, rather than reflecting how the agent itself internally judges its action. Our stand is that the agent's declarative assessments for the state already contain judgment of the possible outcome of each candidate action, and such assessments can be bootstrapped in the same way as the numerical $Q$-function. This is analogous to having a \textit{language value function} which was first proposed by \citep{feng2024natural}, and \textbf{Stateful Knowledge Learning} can be seen as a more general extension of it.

\section{Conclusions and limitation}
\vspace{-0.2cm}
In this paper, we have challenged the prevailing reliance on trajectory-level reflection in LLM agent training. To address this, we introduced \textbf{Stateful Knowledge Learning (SKL)}, a framework that centers on maintaining explicit, declarative predictive assessments anchored to specific states. Our motivating examples confirmed that stateful knowledge provides finer granularity and generalization, and also enables knowledge bootstrapping that scales SKL to increasingly complex tasks. Experiments validate how  SKL training variants \textbf{SKL-SD} and \textbf{SKL-RL} can yield significant performance gains and move toward a more autonomous, self-evolving agent learning paradigm.

As LLM agents transition further into the "Era of Experience," moving beyond static datasets toward continuous environmental interaction, the ability to maintain a stateful "mental map" of predictive foresight will be critical. While our current implementation introduces a trade-off in training compute, the resulting gains in policy robustness and strategic depth suggest that stateful learning is a more principled foundation for autonomous agents. Future work will explore the practical implementation to resolve the computational limitation raised in the paper and the potential for transferring stateful predictive knowledge across heterogeneous task domains.

\bibliographystyle{plainnat}
\bibliography{main}

\clearpage
\appendix

\appendix

\clearpage

\section{Algorithm pseudocode}\label{appendix: algo pseudocode}

\begin{algorithm}[htbp]
  \begin{algorithmic}[1]
    \REQUIRE Policy $\pi_{\theta}$, Environment $E$, Outcome verifier $R$, Bootstrap budget $(N, H)$
    \ENSURE $\pi_{\theta}$
    \STATE Initialize global experience buffer $\mathcal{B}\leftarrow \emptyset$
    \FOR{Epoch $e = 1$ to $K$}
        \STATE Initialize dataset $\mathcal{D}_{\text{base}}, \mathcal{D}_{\text{distill}}, \mathcal{D}_{\text{reflect}} \leftarrow \emptyset,\emptyset,\emptyset$
        \STATE \textcolor{blue}{\textit{\# Base Stateful Rollout}}
        \STATE Sample base stateful trajectories in batches: $\bm{\tau}^e = \{\tau_1, \tau_2, ..., \tau_{N}\} \sim \pi_{\theta_{\text{old}}}$
        \STATE Add transition to global experience buffer $\mathcal{B} \leftarrow \tau^e$
        \FOR{each base trajectory $\tau_i$ in $\bm{\tau}^e$}
            \IF{$R(\tau_i) > 0$}  
                \STATE $\mathcal{D}_{\text{base}} \leftarrow \tau_i$
            \ELSE 
                \STATE \textcolor{blue}{\textit{\# Bootstrapping on failed trajectories}}
                \FOR{timestep $t=T-1$ to $0$}
                    \STATE Sample aggregated trajectories $\{{\tau}_n^H(s_t^{(i)})\}_{n=1}^N \sim \mathcal{B}$ originated at state $s_t^{(i)} \in \tau_i$
                    \STATE Generate updated state knowledge $\hat{z}^{e}_{s_t^{(i)}}\sim \pi_{\theta_{\text{old}}}\big(\cdot \big|s_t^{(i)}, \{{\tau}_n^H(s_t^{(i)})\}_{n=1}^N, z_{s_t^{(i)}}^{e-1}, z^{*}_{s_{t+H}^{(i)}}\big)$
                \ENDFOR
                \STATE \textcolor{blue}{\textit{\# Verify updated state knowledge via Rejection Sampling}}
                \STATE Sampling retried stateful trajectory $\tau^{\text{retry}}_i$ with pre-defined state knowledge $\{\hat{z}^{e}_{s_t^{(i)}}\}_{t=0}^{T-1}$
                \IF{$R(\tau^{\text{retry}}_i) > 0$}
                    \STATE $\mathcal{D}_{\text{distill}}\leftarrow \tau^{\text{retry}}_i$, $\mathcal{D}_{\text{reflect}}\leftarrow \big(\big[s_t^{(i)}, \{\bm{\tau}_n^H(s_t^{(i)})\}_{n=1}^N, z_{s_t^{(i)}}^{e-1}, z^{*}_{s_{t+H}^{(i)}}\big], \hat{z}^{e}_{s_t^{(i)}}\big)$
                \ENDIF
            \ENDIF
            \STATE \textcolor{blue}{\textit{\# Policy Update with SKL-SD Objective Function}}
            \STATE $\mathcal{L}_{\text{SKL-SD}}(\theta) = \mathcal{L}_{\text{PG}}(\mathcal{D}_{\text{base}}) + \lambda_1 \mathcal{L}_{\text{PG}}(\mathcal{D}_{\text{distill}}) + \lambda_2 \mathcal{L}_{\text{SFT}}(\mathcal{D}_{\text{reflect}})$
            \STATE $\theta_{\text{old}}\leftarrow \argmin_{\theta}(\mathcal{L}_{\text{SKL-SD}}(\theta))$
            \ENDFOR 
        \ENDFOR
  \end{algorithmic}
  \caption{\textbf{State-Knowledge-Learning Self-Distillation} (SKL-SD)}\label{algo: sg-sd}
\end{algorithm}

\renewcommand{\thefootnote}{\fnsymbol{footnote}}
\footnotetext[1]{$z_{s_{t+H}^{(i)}}$ can be retrieved from the buffer or the previous time-step in the same bootstrapping loop.}
\renewcommand{\thefootnote}{\arabic{footnote}}

\begin{algorithm}[htbp]
  \begin{algorithmic}[1]
    \REQUIRE Policy $\pi_{\theta}$, Environment $E$, Simulator $E^{\text{sim}}$, Outcome verifier $R$, Bootstrap budget $(N, H)$
    \ENSURE $\pi_{\theta}$
    \FOR{Epoch $e = 1$ to $K$}
        \STATE Initialize dataset $\mathcal{D}_{\text{SKL-RL}}, \mathcal{D}_{\text{distill}} \leftarrow \emptyset,\emptyset$, environmental state $s_t$
        \STATE \textcolor{blue}{\textit{\# Aggregated Stateful Rollout via Simulation}}
        \FOR{$t = 0$ to $T$}
            \STATE Simulate $N$ parallel stateful trajectories $\bm{\tau}_N^H (s_t) = \{\tau_1^{H}(s_t), ..., \tau_{N}^H(s_t)\}$ with horizon $H$
            \STATE Generate updated state knowledge $\hat{z}_{s_t}\sim \pi_{\theta_{\text{old}}}(\cdot | s_t, z_{s_t}, \bm{\tau}_N^H (s_t))$
            \STATE Generate exploiting action $a_t \sim \pi_{\theta_{\text{old}}}(\cdot|s_t, \hat{z}_{s_t})$
            \STATE Transition $s_{t+1} \sim E(s_t, a_t)$, $s_{t} \leftarrow s_{t+1}$
            \ENDFOR 
            \STATE Collect trajectories $\bm{\tau}^e=[s_0, (z_{s_0}, \bm{\tau}_N^H (s_0), \hat{z}_{s_0}, a_0), s_1, (z_{s_1}, \bm{\tau}_N^H (s_1), \hat{z}_{s_1}, a_1), ..., s_T]$
            \STATE $\mathcal{D}_{\text{SKL-RL}}\leftarrow \bm{\tau}^e$, $\mathcal{D}_{\text{distill}}\leftarrow \{(z_{s_t}, \hat{z}_{s_t})\}_{t=0}^{T-1}$
            \STATE \textcolor{blue}{\textit{\# Policy Update with SKL-RL Objective Function}}
            \STATE $\mathcal{L}_{\text{SKL-RL}}(\theta) = \mathcal{L}_{\text{GRPO}}(\mathcal{D}_{\text{SKL-RL}}) + \lambda \mathcal{L}_{\text{Distill}}(\mathcal{D}_{\text{distill}})$\;, \;\;$\theta_{\text{old}} \leftarrow \argmin_{\theta}(\mathcal{L}_{\text{SKL-RL}}(\theta))$
        \ENDFOR
  \end{algorithmic}
  \caption{\textbf{State-Knowledge-Learning Reinforcement Learning} (SKL-RL)}\label{algo:sg-rl}
\end{algorithm}


\section{Notation}

\begin{table}[H]
\centering
\renewcommand{\arraystretch}{1.3}
\begin{tabular}{p{0.14\linewidth} p{0.8\linewidth}}
\toprule
\textbf{Symbol} & \textbf{Description} \\
\midrule
$E$ & Environment \\
$E^{\text{sim}}$ & A resettable environmental simulator \\
$\mathcal{S}$ & Environmental state space \\
$s_t$ & Environmental state at time $t$ \\
$\pi_{\theta}$ & Policy function with parameter $\theta$ \\
$\mathcal{A}$ & Action space \\
$a_t$ & Executable action at time $t$ \\
$z_{s_t}$ & Generated state knowledge at state $s_t$ \\
$\{z_s\}_{s\in\mathcal{S}}$ & State knowledge table \\
$z(\cdot)$ & Generation function for state knowledge \\
$\hat{z}_{s_t}$ & Updated state knowledge at state $s_t$ \\
$(H, N)$ & Bootstrap budget for state aggregation; $H$ is the maximum horizon, $N$ is the number of rollout branches \\
$\tau(s_0)$ & A complete rollout trajectory starting at initial state $s_0$ \\
$\{\tau_n^H(s_t)\}_{n=1}^N$ & $N$ number of sampled partial rollout trajectories originated at $s_t$ with budget $(H,N)$, used in SKL-SD offline bootstrapping stage \\
$\bm{\tau}_{N}^H(s_t)$ & A set online model-based rollout trajectories originated at $s_t$ with budget $(H,N)$, used in SKL-RL online interaction stage \\
$\mathcal{B}$ & Experience buffer \\
$R, R(\cdot)$ & Outcome reward (function) \\
\bottomrule
\end{tabular}
\vspace{5pt}
\caption{Notation Description}\label{tab: notation}
\end{table}

\section{Compare SKL-SD and SKL-RL to other training methods}\label{appendix: compare to other training}

The \textbf{Stateful Knowledge Learning} framework aligns with the \textit{Learning from Language Feedback}~\citep{cheng2023llf} paradigm, as our bootstrapping loop is essentially updating state knowledge using evaluative language generated at subsequent states. The central insight is that \textbf{language feedback conveys richer semantic information than scalar rewards}~\citep{feng2024natural}, regardless of whether it comes from external environments, self-reflection, other models, or human supervision. Within this paradigm, we here compare our methods to recent training-based methods designed to improve self-reflection capabilities. 


Critic-GRPO~\citep{zhang2025critique}, Reflect-Retry-Reward~\citep{bensal2025reflect}, and $\text{R}^3$L~\citep{shi2026r} explore how evaluative behaviour evolves under reinforcement learning. These approaches construct datasets consisting of initial experiences $\mathcal{D}_{\text{base}}$, self-generated reflections $\{d_{\text{reflection}}\}$, and refined experiences $\mathcal{D}_{\text{new}}$ (trajectory-level) or retried experiences $\mathcal{D}_{\text{retry}}$ (step-level), followed by RL training on the combined dataset $\mathcal{D} = \{\mathcal{D}_{\text{base}}, d_{\text{reflect}}, \mathcal{D}_{\text{new/retry}}\}$. They also synthesize reflective reasoning paths by concatenating base-level exploration traces with filtered refinement, denoted as $\{\mathcal{D}_{\text{reflect}}^{\text{distill}}, \mathcal{D}_{\text{retry}}^{\text{distill}}\}$. Table~\ref{table:reflection-RL policy gradient} lists the typical training objectives (policy gradient losses) used for different components. Critic-GRPO \citep{zhang2025critique} can be seen as reinforcing with objectives 1\&2. Reflect-Retry-Reward~\citep{bensal2025reflect} reinforces on objective 6, $\text{R}^3$L \citep{shi2026r} is trained with RL on objective 1\&3, and perform SFT on objective 4\&5. Experiential RL \citep{shi2026experientialreinforcementlearning} optimize with respect to objective 2\&4. 

In our work, \textbf{State-Knowledge-Learning Self-Distillation} (see Equation~\ref{eq: sg-sd}) adopts a similar loss design, but all components follow a \textit{stateful} formulation where state knowledge~$z$ is generated before the action~$a$. Specifically, we maintain a policy gradient loss over base rollouts $\mathcal{L}_{\text{PG}}(\mathcal{D}_{\text{base}})$ (analogous to objective~1, but with stateful trajectories and sample filtering), a policy gradient loss over distilled refined responses $\mathcal{L}_{\text{PG}}(\mathcal{D}_{\text{distill}})$ (analogous to objective~2, again in a stateful form and filtered), and an auxiliary SFT loss $\mathcal{L}_{\text{SFT}}(\mathcal{D}_{\text{reflect}})$ that guides the bootstrapping behaviour. Thus, \textbf{SKL-SD can be viewed as a ``stateful'' version of recent self-reflection RL training methods}; we call it ``self-distillation'' or ``self-imitation'' because we reinforce only filtered (positive) trajectories and discard negative samples.

In contrast, \textbf{State-Knowledge-Learning Reinforcement Learning} (see Equation~\ref{eq: sg-rl}) is \textbf{fundamentally different}: it integrates state aggregation and the bootstrapping loop into a single online rollout episode and relies solely on the final outcome reward to jointly shape both generation and evaluation behavior. This makes it a more unified and autonomous agent training framework.



\begin{table}[htbp]
\renewcommand{\arraystretch}{1.5}
\centering
\small
\begin{tabular}{|p{4.2cm}|p{8.8cm}|}
\hline
\textbf{Objective} & \textbf{Loss Function} (with or without sample filtering) \\
\hline

1. {Base Response} &
$\mathbb{E}_{\mathcal{D}_{\text{base}}}
\left[
R_{\text{base}} \cdot
\nabla_{\theta}\log \pi_{\theta}(\tau_{\text{base}} \mid s_0)
\right]$  \\
\hline

2. {Distilled Refined Response} &
$\mathbb{E}_{\mathcal{D}_{\text{reflect}}^{\text{distill}}}
\left[
R_{\text{reflect}} \cdot
\nabla_{\theta}\log \pi_{\theta}(\tau_{\text{new}} \mid s_0)
\right]$  \\
\hline

3. {Distilled Retried Response} &
$\mathbb{E}_{\mathcal{D}_{\text{retry}}^{\text{distill}}}
\left[
R_{\text{retried}} \cdot
\nabla_{\theta}\log \pi_{\theta}(\tau_{\text{base}, <\text{pivot}}, \tau_{\text{retry}, >\text{pivot}} \mid s_0)
\right]$  \\
\hline

4. {Complete Refinement Process} &
$\mathbb{E}_{\mathcal{D}_{\text{reflect}}}
\left[
R_{\text{reflect}} \cdot
\nabla_{\theta} \log \pi_{\theta}([\tau_{\text{base}}, d_{\text{reflection}}, \tau_{\text{new}}] \mid s_0)
\right]$ \\
\hline

5. {Complete Retry Process} &
$\mathbb{E}_{\mathcal{D}_{\text{retry}}}
\left[
R_{\text{retry}} \cdot
\nabla_{\theta}\log \pi_{\theta}([\tau_{\text{base}}, d_{\text{reflection}}, [\tau_{\text{base}, <\text{pivot}}, \tau_{\text{retry}, >\text{pivot}}]] \mid s_0)
\right]$  \\
\hline

6. {Reflection} &
$\mathbb{E}_{\{d_{\text{reflection}}\}}
\left[
R_{\text{reflect}} \cdot
\nabla_{\theta} \log \pi_{\theta}(d_{\text{reflection}} \mid s_t, \tau_{\text{base}})
\right]$  \\
\hline



\end{tabular}
\vspace{0.3cm}
\caption{Training objective for different purposes in works related to other self-reflection training methods.}
\label{table:reflection-RL policy gradient}
\end{table}

\section{Localized parallel simulation in SKL-RL}\label{appendix: local parallel simulation}

In the context of SKL-RL, the localized parallel simulation mechanism acts as an online, text-based look-ahead tree exploration. Because the agent has access to a resettable simulator, it doesn't just pick an action and move forward blindly; instead, it pauses at the current state $s_t$, freezes the main trajectory, and "probes" the future. Here is a detailed breakdown of how this parallel simulation operates at each timestep:
\begin{enumerate}
    \item \textbf{State Anchoring} (The Root): At any given online step, the agent encounters state $s_t$. The simulator's state is checkpointed or saved at this exact position, acting as the root node for the simulations.
    \item \textbf{Multi-Branch Spawning} (Width $N$): From this single root state $s_t$, the agent generates knowledge $z_{s_t}$, and instantiates $N$ actions for independent, parallel branches (or simulation workers).
    \item \textbf{Independent Temporal Expansion} (Horizon $H$): At each subsequent simulation timestep $1, \dots, H$, every branch rolls out independently. Each branch samples knowledge and actions according to the agent's current policy (or an exploration policy) and receives independent observations and rewards from its own isolated instance of the environment. 
    \item \textbf{Experience Aggregation}: Once all $N$ branches reach the maximum simulation horizon $H$ (or hit a terminal state), their complete rollout histories are gathered into an aggregated experience block $\bm{\tau}_N^H$.
\end{enumerate}
A concrete example is also provided in Figure~\ref{fig:nlts chess example}.

\section{Multi-turn self-distillation in SKL-RL}\label{appendix: distill term}
The primary objective of the distillation loss $\mathcal{L}_{\text{Distill}}(\{z_{s}, \hat{z}_{s}\}_{s\in\mathcal{S}})$ is to internalize the insights gained from look-ahead simulations into the agent's prior stateful knowledge. By grounding its reasoning in previously distilled knowledge, the agent can more effectively explore alternative trajectories in subsequent iterations and incrementally compound its predictive understanding. Formally, this iterative refinement process across consecutive iterations can be expressed as:
\begin{align*}
\text{Iteration } i: & ... \rightarrow s_t \rightarrow \{z_{s_t}, a_t, s_{t+1}, z_{s_{t+1}}, ..., s_{t+H}\} \rightarrow \underline{\hat{z}_{s_t}} \rightarrow a_t \rightarrow s_{t+1} \rightarrow ... \\
\text{Iteration } i+1: & ... \rightarrow s_t \rightarrow \{\underline{\hat{z}_{s_t}}, a_t^{'}, s_{t+1}^{'}, z_{s_{t+1}^{'}}, ..., s^{'}_{t+H}\} \rightarrow \hat{z}_{s_t}^{'} \rightarrow a^{'}_t \rightarrow s_{t+1}^{'} \rightarrow ... 
\end{align*}

This process constitutes a teacher-student framework. The "student" model represents the agent's zero-shot capability to generate stateful knowledge upon first encountering a state, governed by the distribution $\pi(z_{s_t}|s_t, \mathcal{H})$, where $\mathcal{H}$ denotes the interaction history context. The "teacher" model represents the target distribution $\pi(\hat{z}_{s_t}|\bm{\tau}_{N}^H, s_t, \mathcal{H})$, which produces the refined stateful knowledge conditioned on the outcomes of the look-ahead simulations $\bm{\tau}_{N}^H$.The goal of this self-distillation is to align the student model with the teacher, compelling the agent to consolidate simulation-derived experience and autonomously build new knowledge. This alignment can be optimized using either Forward or Reverse Kullback-Leibler (KL) divergence in a multi-turn setting:
\begin{equation}
\textbf{Forward KL}:  \mathcal{L}_{\text{Distill}}^{\text{Foward}}(\theta) = \sum_{s_t} \text{KL}\bigg[\pi_{\theta}(z|\bm{\tau}_{N}^H, s_t, \mathcal{H}) \bigg\| \pi_{\theta}(z|s_t, \mathcal{H}) \bigg]
\end{equation}
\begin{equation}
    \textbf{Reverse KL}: \mathcal{L}_{\text{Distill}}^{\text{Reverse}}(\theta) = \sum_{s_t} \text{KL}\bigg[\pi_{\theta}(z|s_t, \mathcal{H})\bigg\| \pi_{\theta}(z|\bm{\tau}_{N}^H, s_t, \mathcal{H})  \bigg]
\end{equation}

Here, the Forward KL objective serves as a multi-turn, self-distillation extension of Supervised-KD \citep{agarwal2024policy}, while the Reverse KL acts as a multi-turn extension of OPSD \citep{zhao2026self}.

\section{Experiments on agentic environments}

\subsection{Experiment setup}\label{appendix: skl-sd exp setup}
The experimental setup of SKL-SD on agentic tasks is highly aligned with the setup used by \citep{shi2026r} as we re-implement SKL-SD in their public codebase, and use the exact hyperparameter setting reported in their paper. The baseline includes rejection sampling methods like RAFT \citep{dong2023raft}, GRPO variants including the default GRPO \citep{guo2025deepseek}, OPMD \citep{yao2025group}, and GSPO \citep{zheng2025group}. Crucially, we compare against language-feedback frameworks that utilize self-reflection or critiques, including Critique-GRPO \citep{zhang2025critique}, Reflect-GRPO \citep{bensal2025reflect}, and R$^3$L \citep{shi2026r}. All experiments are implemented using the Trinity-RFT framework \citep{pan2025trinity} to ensure parity in infrastructure. We use \textit{Qwen2.5-7B-Instruct} \citep{qwen2} as the backbone model across all methods. The training is operated on a NVIDIA H100 single-node cluster.

\subsection{Detailed reasoning trace}\label{appendix: webshop skl-sd trace}

\begin{tcolorbox}[
  breakable,
  colback=grey!5!white,
  colframe=grey!50!white,
  code={\refstepcounter{table}}, 
  title={Table \thetable: A reasoning example on WebShop. Multiple state knowledge revealed at different states},
  label={table: webshop skl-sd trace}
]

\footnotesize

\textbf{Query}:

Find me non-slip, easy clean computer armoires with PU leather with color: white, and size: 120x60cm, and price lower than 70.00 dollars \medskip
\hrule
\medskip

\textcolor{blue}{\textbf{[State Knowledge for Initial Search]}}

\textbf{State Understanding}:

1. Search Queries:
\begin{itemize}
    \item Ensure the search terms include specific attributes and measurements to match the exact product requirements:     - Color: … 
    - Size: …    
    - Additional Features:…    
    - Price: …
\end{itemize}

2. Product Verification and Filtering:
\begin{itemize}
    \item After locating a potential product, verify all criteria (color, size, features, price) are correctly specified.
    \item Directly click "Buy Now" if the product matches all specified criteria, without additional checks for availability.
\end{itemize}

3. Buy Now on Detailed Product Page:
\begin{itemize}
    \item Ensure that the click on the "Buy Now" button is performed directly on the detailed product page with all filters correctly applied.
    \item Click the "Buy Now" button if all conditions are verified.
\end{itemize}

4. Detailed Review and Double-Check:
\begin{itemize}
    \item Verify that the product's page correctly displays all the required attributes and features, using inspection tools if necessary.
    \item Double-check that the product's details, especially dimensions (size: 120x60cm) and color (white), and price ($\leq$ \$70.00) match the criteria.
\end{itemize}

5. Error Logging:
\begin{itemize}
    \item If a product does not match all criteria, log the relevant details of the product and try again, ensuring a revisit to the search page with corrected parameters if necessary.
\end{itemize}

\textbf{Steps for Implementation:}

1. Refine Search Query with Specific Attributes and Dimensions: ……

\nopagebreak\medskip
\noindent{\color{black!50!white}\tikz\draw[dash pattern=on 3pt off 2pt, line width=1.5pt] (0,0) -- (\linewidth,0);}
\nopagebreak\medskip

\textcolor{blue}{\textbf{[State Knowledge for Browsing]}}

\textbf{State Understanding:}

1. Correct Product Matching: Ensure that the selected product details (color, size, price)—all attributes must correctly match the required criteria: 1. Color: ..., 2. Size: ..., 3. Price: ...

2. Product Detail Verification:
\begin{itemize}
    \item Double-check each product detail on the result page to ensure accuracy.
    \item Verify that images and descriptions explicitly confirm the exact details being used (white color, 120x60cm size, etc).
\end{itemize} 

......

N. Notes for Future Runs:
\begin{itemize}
    \item Cross-Verify Attributes: Validate the exact color, size, and price on each detailed product page.
    \item Double-check Final Product: If unsure, re-verify the product details and attributes before clicking Buy Now.
    \item Snapshots as Evidence: Take screenshots or notes of the product description and attributes to confirm they meet the criteria.
\end{itemize}

\nopagebreak\medskip
\noindent{\color{black!50!white}\tikz\draw[dash pattern=on 3pt off 2pt, line width=1.5pt] (0,0) -- (\linewidth,0);}
\nopagebreak\medskip

\textcolor{blue}{\textbf{[State Knowledge for Specific Product Page]}}

\textbf{State Understanding}:

1. Navigating the URL: Start by ensuring that the correct URL is accessed, specifically B095KCTB3Y, to display the desired product.

2. Selecting Size and Color Details:
\begin{itemize}
    \item Ensure that the size 120x60cm and color white are explicitly selected and chosen correctly from dropdown options (if available).
    \item Double-check that the product details on the page reflect the correct options and attributes (white, 120x60cm, non-slip).
\end{itemize}

3. Direct Purchase Action: Once all conditions (size and color) are correctly matched and verified on the product description, click the Buy Now button.

\textbf{Notes}:
\begin{itemize}
    \item Ensure that the search query uses keywords such as "white", "120x60cm", and "pu leather" to precisely target the correct product.
    \item Directly interact with the specified URL and verify the exact attributes before finalizing the purchase.
    \item Use the correct Buy Now button to conclude the task.
    \item  For future iterations, focus on refining the search criteria to eliminate confusion.
\end{itemize}

\end{tcolorbox}

\section{Experiments on ChessPuzzles}\label{appendix: exp chess}

\subsection{The uniqueness of the chess game for LLM evaluation}\label{appendix: motivation for chess} 
Recent research on LLM-based agents often evaluates performance on relatively simple agentic environments, such as ALFWorld \citep{Shridhar2020ALFWorldAT}, WebShop \citep{Yao2022WebShopTS}, or grid-based games like FrozenLake and Sokoban \citep{feng2024natural}. While these benchmarks are useful for studying sequential decision-making and tool-use behaviors, they typically involve limited strategic depth and can often be solved through short-horizon reasoning or heuristic exploration. Consequently, many agent studies primarily report final task completion rates, without closely examining the soundness or internal consistency of the reasoning traces that lead to those outcomes. In contrast, chess games require substantially richer strategic reasoning \citep{liu2025chessarena}. Solving a chess puzzle generally involves multi-step planning, evaluation of alternative move sequences, and understanding tactical motifs such as checks, pins, or forced mates. These characteristics make chess a more demanding testbed for assessing whether an agent can produce coherent reasoning processes rather than merely reaching the correct final answer.

Another advantage of chess puzzles is that they are less susceptible to data contamination compared to many popular agent benchmarks. Many environments have well-documented task templates that may appear in pretraining corpora, making it difficult to disentangle memorization from genuine reasoning. In contrast, our experiments indicate that even strong closed-source models (e.g., GPT-5) struggle to solve the ChessPuzzles benchmark in a zero-shot setting, as illustrated in  Table~\ref{table: chess bad reasoning examples}. This suggests that the model does not simply recall solutions from training data but must instead need to reason about the game dynamics.

Taken together, these properties make chess puzzles an ideal evaluation setting for studying an agent’s ability to adapt to an environment with minimal prior knowledge of the underlying rules. This motivates our use of chess as a challenging benchmark to evaluate the effectiveness of our algorithm.

\begin{tcolorbox}[
  breakable,
  colback=grey!5!white,
  colframe=grey!50!white,
  code={\refstepcounter{table}}, 
  title={Table \thetable: Reasoning of based model in ChessPuzzles tasks. Both Qwen2.5-7B-Instruct and GPT-5 fail to provide good reasoning.},
  label={table: chess bad reasoning examples}
]

\footnotesize
\textbf{Game State: 1Qr5/p4r1k/8/3p2p1/P2P2P1/2P2pKP/8/8 w - - 0 49} \;,\;\; \textbf{StockFish's Move: Qb1}
\medskip
\hrule
\medskip

\noindent\textbf{Qwen2.5-7B-Instruct} \textcolor{red}{(\textbf{Nonsensical reasoning})} \\[4pt]
\scriptsize
Let's analyze the position strategically. White has a queen and a king, while Black has a rook and a king. The position is closed, with few pieces on the board, so tactical play is important.\\
\\
\textbf{Key candidate moves to consider are}:

1. Qxc8: Captures the pawn and puts pressure on the black king. \textcolor{red}{(\textbf{Misidentified the piece})}\\
2. Qa8: Attacks the rook on a1 and also threatens the black king. \textcolor{red}{(\textbf{Misidentified the piece})} \\
3. Qc7: Targets the black king directly. \textcolor{red}{(\textbf{Wrong justification})}\\
\\
Now let's \textbf{evaluate each move}: \textcolor{red}{(\textbf{Completely wrong})}\\
1. \textbf{Qxc8}: If this move is played, Black will respond with Rxa8. The resulting position will leave White with a potential fork with the queen, targeting both the black king and the rook. However, this might be risky as Black could still create threats with the rook on a8.\\
2. \textbf{Qa8}: This move attacks the rook on a1 and threatens to put pressure on the black king. After Qa8, Black can respond with Rxa8, but this would result in a simplified position where White can focus on controlling the center and potentially queening the pawn on h4 or a5.\\
3. \textbf{Qc7}: This move directly targets the black king, putting it under pressure. If Qc7 is played, Black can respond with Kf8, which limits the immediate threat. However, it leaves the black king more exposed.\\
\\
Considering the strategic implications, Qa8 seems to be the most promising move. It not only attacks a key piece but also creates a significant threat to the black king. The resulting position is simpler and allows White to control the center and potentially create winning chances. \texttt{<move>}Qa8\texttt{</move>}

\vfill
\tcblower 

\noindent\textbf{GPT-5} (\textcolor{ForestGreen}{\textbf{Better at general reasoning}}, \textcolor{red}{\textbf{but fail at detailed analysis}}) \\[4pt]
\scriptsize
\textbf{Strategic evaluation} \textcolor{ForestGreen}{(\textbf{Making Sense})}:\\
\\
Material is heavily in White’s favor because White has a queen (on b8) against Black’s rook (on f7) plus some pawns. The immediate priority is to simplify into a trivially winning endgame by trading queens/major pieces or winning the opposing rook, while also keeping White’s king safe. Black’s king is on h7 and is relatively exposed; Black’s main counterplay is the rook on f7 plus the advanced f3-pawn.\\
\\
\textbf{Three key candidate moves}:\\
\\
1) Qxc8\\
Idea: capture the black queen on c8 and trade queens immediately.\\
Likely response: Black must recapture with Rxc8 (the rook from f8 goes to c8).\\
Outcome: White has traded queens, but then White is only up material modestly and still has to deal with Black’s active rook and passed f-pawn. This is good and likely winning, but it gives Black some practical chances compared to winning the rook outright. \textcolor{red}{(\textbf{Wrong detailed analysis})}\\
\\
2) Qxa7\\
Idea: win a pawn with check possibilities later (queen near the king).\\
Likely response: Black can continue with f2 (or Rxa7 isn’t possible), pushing the passed pawn and creating real promotion threats, while White hasn’t reduced Black’s major-piece activity.\\
Outcome: This is slower and allows Black counterplay with the f-pawn and rook coordination. Not the cleanest conversion.\\
\\
3) Qb7\\
Idea: directly attack Black’s rook on f7 along the 7th rank and win it.\\
Likely response: Black’s rook has no squares available on the 7th rank because b7 controls the entire rank; it also cannot capture Qb7 (rook on f7 can’t move like that) \textcolor{red}{(\textbf{Wrong prediction})}, and moving the king doesn’t solve the attack. The only “defense” would be moving the rook away—but from f7, every legal rook move is along the f-file or 7th rank, and Qb7 covers f7 and the whole 7th, so the rook is effectively lost.\\
Outcome: White wins the rook cleanly, after which the position is completely winning with queen versus only pawns.\\
\\
\textbf{Conclusion}:\\
Qb7 is the most forcing and highest-value move: it wins Black’s rook immediately with minimal counterplay, converting the advantage far more cleanly than trading queens with Qxc8 or grabbing a pawn with Qxa7. \texttt{<move>}Qb7\texttt{</move>}

\end{tcolorbox}

\subsection{Experiment Setup}\label{appendix: chesspuzzle other setup}

\paragraph{Experiment Setup} We also compare SKL-RL with other training baselines introduced in Section~\ref{section: sg-sd exp}, all of which operate via trajectory-level self-reflection. All methods are trained with matched computation budgets, measured by the number of LLM calls during training. We use Qwen2.5-7B-Instruct as the base model, and apply GRPO for all RL implementations. We select a subset of the original ChessPuzzles (rating $<1000$) and split it into a train and a test set.

\begin{figure}[htbp]
    \centering
    \includegraphics[width=0.8\linewidth]{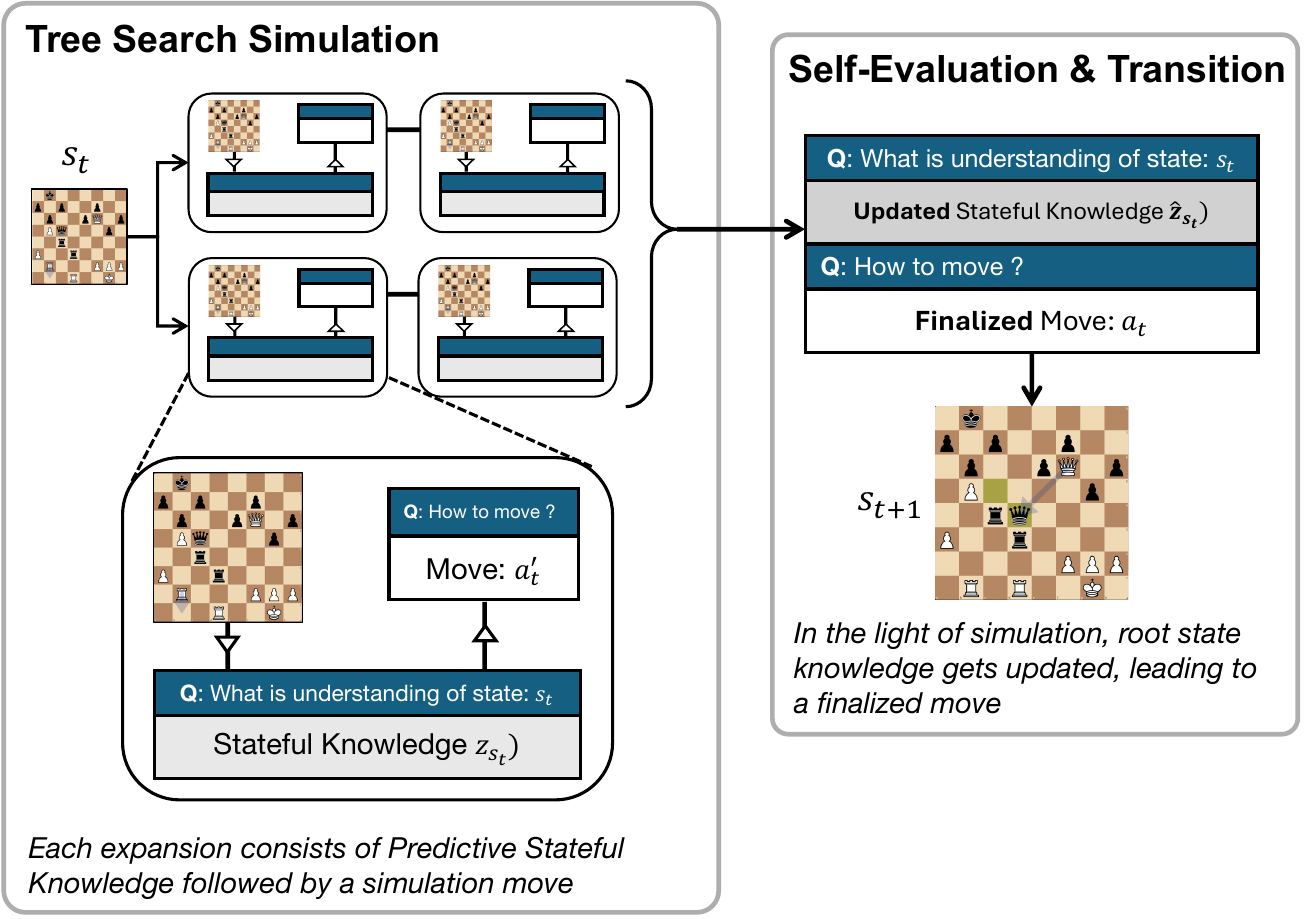}
    \caption{At each state $s_t$, the agent first performs a tree search where each node expansion consists of state knowledge and a simulation move. The generated simulation experience $\bm{\tau}_N^H(s_t)$ is aggregated as the context, and the agent summarizes an updated state understanding $\hat{z}_{s_t}\sim \pi_{\theta}(\cdot|\bm{\tau}_N^H(s_t), s_t, z_{s_t})$ as well as a finalized move $a_t$ to transition to the next state $s_{t+1}$.}
    \label{fig:nlts chess example}
\end{figure}

On ChessPuzzles, agents are prompted to generate state knowledge $z(s_t)$ for the current game state $s_t$ and decide on the movement $a_t$ to make in two consecutive reasoning steps:
\begin{enumerate}
    \item \textbf{State Knowledge} $z_t$: \texttt{[User]: What is your understanding of the game state: \{state\}. [Assistant]: ...}
    \item \textbf{Movement} $a_t$: \texttt{[User]: You are in stage \{stage\}, based on the state understanding, what is the decided move? [Assistant]: ...}
\end{enumerate}
where the state knowledge explicitly queries the model for its internal predictive knowledge of the game state, and the movement query will use the generated understanding as the context, asking the model to directly generate action at each stage (i.e., both during simulation or bootstrapping). This way of reasoning can also avoid the issue of overthinking that causes missing action tokens and make knowledge distillation easier to implement later. Figure~\ref{fig:nlts chess example} illustrates this. Self-generated tree search simulation is carried out in a similar way as parallel decoding, where the agent is prompted to generate multiple ($H$) moves at the same time, and for each rollout branch, the agent independently performs a single-trajectory rollout until running out of horizon budget. Table~\ref{table: chess system_prompt} shows the system prompt for ChessPuzzles.

\begin{tcolorbox}[
  breakable,
  colback=grey!5!white,
  colframe=grey!50!white,
  code={\refstepcounter{table}}, 
  title={Table \thetable: System Prompt used in ChessPuzzles tasks.},
  label={table: chess system_prompt}
]

\footnotesize

A conversation between User and Assistant. \\
The User asks the best move to make for a given chess board state, and the Assistant solves it. \\
The Assistant is a professional chess player who can plan a few steps ahead by interacting with a simulator, and then think about the reasoning process and at last provides the User with the answer.\\
\\
The Assistant's thinking process must include an understanding of the game state, based on which the Assistant can later decide on the next move.\\
The understanding should describe how the Assistant analyzes the current position, including:\\
- A strategic evaluation of the position.\\
- A comparison of key candidate moves (not all of them) in terms of their possible subsequent outcomes.\\
- Do not try to enumerate all possible moves and outcomes, only consider the most likely ones.\\
\\
The decision on next move must be in SAN notation, strictly using the moving piece and the destination square (e.g., Nf3, Rxf2, c5).\\
\\
Reminder of chess rules: \\
- Bishops move diagonally. \\
- Rooks move horizontally or vertically. \\ 
- Knights jump in an L-shape.  \\
- Queens combine rook and bishop movement. \\
- Kings move one square in any direction. \\
- Pawns move forward, capture diagonally, and can promote.\\
\end{tcolorbox}

\subsection{Ablation on bootstrapping behaviour at different heuristic simulation strategies}\label{appendix: exp sg-rl heuristic reasoning trace}

\begin{tcolorbox}[
  breakable,
  colback=grey!5!white,
  colframe=grey!50!white,
  listing options={
    basicstyle=\ttfamily\footnotesize,
    breaklines=true,
    breakatwhitespace=false,
    columns=fullflexible,
  },
  code={\refstepcounter{table}}, 
  title={Table \thetable: Reasoning trace example of ablation study for SKL-RL in Section~\ref{section: exp sg-rl heuristic ablation}.},
  label={table: chess sg-rl heuristic ablation reasoning trace}
]

\footnotesize

\noindent\textbf{(Test set) Game State: 2r3k1/p1r2ppp/2N1p3/3pPP2/4b1P1/1P6/P6P/2R2RK1 w - - 1 28}
\medskip
\hrule
\medskip

\noindent\textbf{Simulation Strategy: Expert Move Rate} $\frac{1}{2}$ in simulation \textcolor{blue}{(\textbf{short, frequency-based})}
\smallskip
\nopagebreak\medskip{\color{grey!30!white}\hrule}\nopagebreak\medskip
\medskip

\scriptsize
The given game state is: 2r3k1/p1r2ppp/2N1p3/3pPP2/4b1P1/1P6/P6P/2R2RK1 w - - 1 28.
From the simulation trajectories, we can see that the move `Nd8` is the most frequently chosen move by the player, and it leads to a few different successor states. Let's analyze the trajectories:

\smallskip
\texttt{[Simulation Trajectory]}
\smallskip

From the trajectories, we can see that the move `Nd8` leads to the same or similar states in all cases where it is played. On the other hand, the move `Ne7` leads to a different state. The state after `Nd8` is: 2rN3k/p1r2ppp/4p3/3pPP2/4b1P1/1P6/P6P/2R2RK1

This state is very similar to the original state, suggesting that `Nd8` is a good move as it maintains the position quality. The state after `Ne7` is: 2r4k/p1r1Nppp/4p3/3pPP2/4b1P1/1P6/P6P/2R2RK1

This state is different and potentially less favorable, as it involves a knight move that could be less effective in maintaining the position.

\smallskip
\textbf{Strategic evaluation}:
\begin{itemize}[leftmargin=*,noitemsep,topsep=2pt]
  \item The position is balanced, with both sides having similar material and pawn structures.
  \item The move `Nd8` supports the king and maintains the structure, while `Ne7` leads to a different and potentially less favorable position.
\end{itemize}

Based on the simulation rollouts and the strategic evaluation, the best move is: \textbf{Nd8}
\vfill
\nopagebreak\medskip
\noindent{\color{black!50!white}\tikz\draw[dash pattern=on 3pt off 2pt, line width=1.5pt] (0,0) -- (\linewidth,0);}
\nopagebreak\medskip

\footnotesize
\noindent\textbf{Simulation Strategy: Expert Move Rate} $\frac{1}{6}$ in simulation \textcolor{blue}{(\textbf{more structured, but hallucinate})}
\smallskip
\nopagebreak\medskip{\color{black!20!white}\hrule}\nopagebreak\medskip
\medskip
\scriptsize

The given state is: 2r3k1/p1r2ppp2N1p3/3pPP24b1P1/1P6P6P/2R2RK1w - -1 28. This is a middlegame position with White to move. Let's analyze the key features of this position:

\begin{enumerate}[leftmargin=*,noitemsep,topsep=2pt]
  \item \textbf{Material Balance}: The material is balanced with two rooks (R), two knights (N), and one bishop (B) for each side. White has a pawn (P) on the 8th rank (White's king-side), and Black has a pawn on the 7th rank (Black's king-side).
  \item \textbf{King Safety}: Both kings are castled on the kingside, and neither side has any immediate threats to their kings.
  \item \textbf{Pawn Structure}: White has a pawn on e2 and pawns on f2 and g2. Black has pawns on e7 and f7.
  \item \textbf{Active Pieces}: White has a knight on b8 and a rook on the 1st rank. Black has a knight on a5 and a rook on the 1st rank.
  \item \textbf{Key Moves}: The knight on b8 (Nb8) in Traj 3 suggests that it might be a strong move to support the pawn structure and potentially create threats. The knight moves Ne7 and Nb5 in Traj 1 and 2 suggest support for the king-side pawns and maintaining control over key squares. The rook move Na5 in Traj 4 suggests a support move for the knight and maintaining control over key files.
\end{enumerate}

\textbf{Analysis of Moves}
\begin{itemize}[leftmargin=*,noitemsep,topsep=2pt]
  \item \textbf{Move: Nb8 (Traj 3)}: Supports the pawn structure, maintains control over key squares, and potentially prepares for future pawn moves.
  \item \textbf{Move: Ne7 (Traj 1)}: Supports the king-side pawns and maintains control over key squares.
  \item \textbf{Move: Na5 (Traj 4)}: Supports the knight and maintains control over key files.
\end{itemize}

\textbf{Conclusion}: The move Nb8 in Traj 3 seems to be a strong candidate as it supports the pawn structure and maintains control over key squares. Given the similarity in the trajectories, maintaining support and control over key squares appears to be a strong strategy.

\textbf{Recommendation}: The best move for White in the given state is: \textbf{Nb8}
\vfill
\nopagebreak\medskip
\noindent{\color{black!50!white}\tikz\draw[dash pattern=on 3pt off 2pt, line width=1.5pt] (0,0) -- (\linewidth,0);}
\nopagebreak\medskip

\footnotesize
\noindent\textbf{Simulation Strategy: Random Move} in simulation \textcolor{blue}{(\textbf{ignore simulation, more explicit reasoning but severe hallucination})}
\smallskip
\nopagebreak\medskip{\color{black!20!white}\hrule}\nopagebreak\medskip
\medskip
\scriptsize

\textbf{Analysis of the Game State}

The root state of interest is: 2r3k1/p1r2ppp/2N1p3/3pPP2/4b1P1/1P6/P6P/2R2RK1 w - - 1 28.

This state is derived from the four trajectories provided. Let's break down the key elements of this position:

\begin{enumerate}[leftmargin=*,noitemsep,topsep=2pt]
  \item \textbf{White to Move (w - - 1 28)}: White is to move, and it is the 28th move of the game.
  \item \textbf{Material Balance}: White has 2 rooks, 3 knights, 1 king, and 1 pawn. Black has 2 rooks, 1 knight, 1 king, and 1 pawn.
  \item \textbf{Positional Evaluation}: White has a rook and a knight, while Black has a rook and a knight. White has a king and a rook, while Black has a king and a rook. White has a pawn on the 6th rank, while Black has a pawn on the 6th rank.
  \item \textbf{Key Pieces}: White's rook on d1 and Black's rook on c4. White's knight on g5 and Black's knight on d8.
  \item \textbf{Active Play}: The game is still in the middlegame, with no clear endgame or decisive advantage. Both sides have active pieces, and the game is likely to continue with tactical exchanges and positional maneuvers.
\end{enumerate}

\textbf{Trajectory Analysis}: Let's analyze the trajectories to understand the potential future value of the root state:
\begin{itemize}[leftmargin=*,noitemsep,topsep=2pt]
  \item \textbf{Traj 1, 2, 3, 4}: The state remains similar, with no clear changes in the position.
\end{itemize}

\textbf{Strategic Evaluation}
\begin{itemize}[leftmargin=*,noitemsep,topsep=2pt]
  \item \textbf{Control of the Board}: Both sides have active pieces, and the board is open.
  \item \textbf{King Safety}: Both kings are relatively safe, but the rooks and knights are more active.
  \item \textbf{Pawn Structure}: The pawns on the 6th rank are key, and both sides have similar pawn structures.
\end{itemize}

\textbf{Candidate Moves}: Given the current state, let's consider the most likely candidate moves:
\begin{enumerate}[leftmargin=*,noitemsep,topsep=2pt]
  \item \textbf{g5}: This move might aim to control the g5 square and potentially create a tactical opportunity.
  \item \textbf{h4}: This move might aim to control the h4 square and potentially create a tactical opportunity.
  \item \textbf{Rfd1}: This move might aim to control the d1 square and potentially create a tactical opportunity.
\end{enumerate}

\textbf{Conclusion}: Based on the analysis of the trajectories, the root state is still in a balanced middlegame position with active pieces. The moves g5, h4, and Rfd1 are all plausible and aim to create tactical opportunities or control key squares.

\textbf{Recommended Move}: Given the strategic evaluation and the potential future value from the trajectories, the move \textbf{g5} seems to be a strong candidate as it controls a key square and potentially creates tactical opportunities. \textbf{Recommended Move: g5}

\end{tcolorbox}

\subsection{A complete example of SKL-RL with self-generated simulation}\label{appendix: sgrl complete trace}

\begin{tcolorbox}[
  breakable,
  colback=grey!5!white,
  colframe=grey!50!white,
  listing options={
    basicstyle=\ttfamily\footnotesize,
    breaklines=true,
    breakatwhitespace=false,
    columns=fullflexible,
  },
  code={\refstepcounter{table}}, 
  title={Table \thetable: Complete reasoning trace of SKL-RL at state $s_t$.},
  label={table: chess sg-rl complete trace}
]

\footnotesize

\noindent\textbf{[System Prompt]}: \\
(See Table~\ref{table: chess system_prompt})
\medskip

\noindent\textbf{[User]} \textcolor{blue}{(\textbf{Environment State} $\mathbf{s}_t^{E}$)} \\
Current Game State:\\
\texttt{k1r4r/p4pp1/2Nb2p1/3P4/Q3p3/3nPq1P/P2K1P2/R4R2 w - - 4 29}
\bigskip

\tcbsubtitle[colback=grey!30!white, colframe=grey!50!white]{Root State Simulation $s_t$}
\medskip

\noindent\textbf{[User]} \textcolor{blue}{(\textbf{Root State $\mathbf{s}_t=s_t^{E}$ in Simulation})} \\
Current Simulation State: \texttt{k1r4r/p4pp1/2Nb2p1/3P4/Q3p3/3nPq1P/P2K1P2/R4R2 w - - 4 29}
\medskip

What is your understanding of the game state: \\
\texttt{k1r4r/p4pp1/2Nb2p1/3P4/Q3p3/3nPq1P/P2K1P2/R4R2 w - - 4 29}
\bigskip

\noindent\textbf{[Assistant]} \textcolor{blue}{(\textbf{Understanding $z_{s_t}$ at Root State})} \\
\textbf{\#\#\# Understanding of the Game State} \\
The current position is \dots
\bigskip

\noindent\textbf{[User]} \\
You are now in simulation, your current goal is to move in order to better understand potential future outcomes of the game, so you can either explore or exploit. You can generate at most 3 \textcolor{blue}{(\textbf{budget} $N$)} candidate moves to be tested with the simulator. Do not generate duplicate moves.

Choose the move from the available move list: \dots
\bigskip

\noindent\textbf{[Assistant]} \textcolor{blue}{(\textbf{Simulation Action $a_t^{(i)}$ at Root State})} \\
\texttt{<move>}Qxa7\texttt{</move>} \ \texttt{<move>}Qa6\texttt{</move>} \ \texttt{<move>}Ne7\texttt{</move>}
\bigskip

\noindent\textbf{[User]} Found 3 valid moves. \textcolor{blue}{(\textbf{Next State $s_{t+1}^{(i)}$ in Simulation})}
\begin{itemize}[leftmargin=*,noitemsep,topsep=2pt]
  \item \textbf{Traj 1}: Subsequent result of move Qxa7: Results: You have successfully solved the game within this simulation!
  \item \textbf{Traj 2}: \dots\dots
  \item \textbf{Traj 3}: \dots\dots
\end{itemize}
\bigskip

\tcbsubtitle[colback=grey!30!white, colframe=grey!50!white]{Expanded State Simulation $s_{t+1}$}
\medskip

\noindent\textbf{[User]} \& \textbf{[Assistant]} \textcolor{blue}{(\textbf{Continue Simulation at $s_{t+1}^{(i)}$})} \\
(Turn 2 simulation) \dots
\bigskip

\tcbsubtitle[colback=grey!30!white, colframe=grey!50!white]{Self-Evaluation/Bootstrapping for $\hat{z}(s_t)$ and $a_t$}
\medskip

\noindent\textbf{[User]} \textcolor{blue}{(\textbf{Simulation Ends, Start Bootstrapping and Get $\hat{z}_{s_t}$})} \\ {}
[Summarizing stage] \\
The simulation ends. Lets pause, check the interaction history, and rethink again both the benefits and harm of each possible moves at simulation step 1. 
\medskip

\textbf{\#\#\# Simulation History}: \dots
\bigskip

\noindent\textbf{[User]} \\
Based on the simulation history, what is your updated understanding of the game state: \texttt{k1r4r/p4pp1/2Nb2p1/3P4/Q3p3/3nPq1P/P2K1P2/R4R2 w - - 4 29}
\bigskip

\noindent\textbf{[Assistant]} \textcolor{blue}{(\textbf{Finalized action $a_t$ at $s_t$})} \\
\textbf{\#\#\# Updated Understanding of the Game State} \dots
\bigskip

\noindent\textbf{[User]} \\
The simulation ends, your current goal is to output the optimal move for the root state. Based on previous simulation, your understanding of the game state and your current goal, what is the best move to make?
\bigskip

\noindent\textbf{[Assistant]} \\
\texttt{<answer>}Qxa7\texttt{</answer>}

\end{tcolorbox}

\subsection{Reasoning trace for R$^3$L and SKL-RL with self-generated simulation after end-to-end training}\label{appendix: skg-rl chess trained trace}

\begin{tcolorbox}[
  breakable,
  colback=grey!5!white,
  colframe=grey!50!white,
  code={\refstepcounter{table}}, 
    title={Table \thetable: Reasoning trace example ONE of SKL-RL after end-to-end RL training. SKL-RL helps with more concise, stateful, outcome-aware reasoning.},
    label={table: chess sg-rl trained trace 1}
]

\scriptsize

\textbf{Game State}: \\
r1b1k2r/5ppp/2p5/p1q1p1B1/Pp2P1n1/1P3NN1/1P3PPP/3Q1RK1 w kq - 0 17 

\medskip
\hrule
\medskip

\textbf{[State Understanding]}\\

The current position is a complex middlegame position for White, who is in a strong position due to their queen and rook being active and threatening the black king. White has a strong initiative and should focus on developing their queenside attack while also preparing to push their queenside pawns. \\

Let's consider the most promising moves:

1. \textbf{Qd8}: This move is very strong because it not only attacks the black queen but also threatens to checkmate on the next move if the queen moves. \textcolor{ForestGreen}{(\textbf{Correctly identify the key benefits})}

2. \textbf{Kh1}: This move is more positional and prepares to push the h-pawns, which could be a long-term strategy.

3. \textbf{Re1}: This move is more defensive but also prepares to support the h-pawns and control the e-file.

4. \textbf{Nxe5}: This move is more tactical and aims to put pressure on the black knight and potentially open up the e5 square for further attacks. \\

\textcolor{ForestGreen}{(\textbf{analysis seems reasonable}} \textcolor{red}{\textbf{but still at a surface-level})}\\

Given the strong positional and tactical threats, the move Qd8 stands out as the most aggressive and decisive. \\


\nopagebreak\medskip
\noindent{\color{black!50!white}\tikz\draw[dash pattern=on 3pt off 2pt, line width=1.5pt] (0,0) -- (\linewidth,0);}
\nopagebreak\medskip

\textbf{[Experience Summary/Bootstrapping]} \\

Based on the simulation history, the updated understanding of the game state is as follows:

1. White is in a strong position, having just won a queen and a rook, which is a significant material advantage. \textcolor{red}{(\textbf{wrong})}

2. White has a strong attack on the black king, with the queen and rook threatening to checkmate on the next move. \textcolor{ForestGreen}{(\textbf{correctly sumarize the key benefits})}

3. The black king is in a vulnerable position, with the queen and rook threatening to deliver checkmate.

4. The black queen is a potential threat, but it is not well-placed to counter the white attack.

5. The white king is safe and can continue to support the attack on the black king.

6. The board is open and dynamic, with many pieces still active and ready to participate in the attack.

Given this updated understanding, the position is a strong position for White, and the move \boxed{\textbf{Qd8}} was the best move in the simulation history, leading to a win. \textcolor{ForestGreen}{(\textbf{combine knowledge with simulation result})}

\end{tcolorbox}

\begin{tcolorbox}[
  breakable,
  colback=grey!5!white,
  colframe=grey!50!white,
  code={\refstepcounter{table}}, 
    title={Table \thetable: Reasoning trace example of R$^3$L. Trajectory reflection from R$^3$L ignores stateful information.},
    label={table: chess r3l trained trace}
]

\scriptsize

\textbf{Game State}: \\
r1b1k2r/5ppp/2p5/p1q1p1B1/Pp2P1n1/1P3NN1/1P3PPP/3Q1RK1 w kq - 0 17

\medskip
\hrule
\medskip

\textbf{[Trajectory Reflection]}\\

\textbf{trajectory summary}: The Assistant analyzed the chess position and decided on a sequence of moves focusing on gaining positional advantage, promoting pawns, and controlling key squares. However, the Assistant failed to complete the task within the step limit, as no pawn promotion occurred despite several opportunities.\\

\textbf{root cause analysis}: Why was the approach suboptimal? Because the Assistant did not prioritize pawn promotion in the key moments. Why did the Assistant not promote pawns? Because the strategy was too positional and did not explicitly plan for pawn promotion. Why this positional approach? Because the Assistant optimized for immediate positional advantages rather than focusing on the critical goal of promoting pawns. Root cause: Lack of explicit focus on the endgame goal of promoting pawns. The Assistant should have immediately promoted a pawn when given the opportunity, as the game was nearing its end. \\

\textbf{trajectory outcome}: failure,\\

\textbf{improvement suggestion}: In endgame scenarios, especially in rook and pawn situations, the primary goal should be to promote pawns. The Assistant must explicitly prioritize pawn promotion and take advantage of opportunities to do so, even if it means sacrificing other positional benefits. For instance, when the King is closer to the fifth rank (more pawn promotion squares), the Assistant should move towards those squares and promote the pawn as soon as possible. Example: In the current trajectory, when the white king was on g7 or h6, the priority should have been to move towards a pawn promotion square and promote the pawn immediately when legal. \\

\textbf{retry from step}: 0    \\
\textcolor{red}{\textbf{(Coarse-grained Understanding)}} \\

\end{tcolorbox}

\begin{tcolorbox}[
  breakable,
  parbox=false,
  colback=grey!5!white,
  colframe=grey!50!white,
  listing options={
    basicstyle=\ttfamily\footnotesize,
    breaklines=true,
    breakatwhitespace=false,
    columns=fullflexible,
  },
    code={\refstepcounter{table}}, 
    title={Table \thetable: Reasoning trace example TWO of SKL-RL before and after end-to-end RL training.},
    label={table: chess sg-rl trained trace 2}
]

\footnotesize

\textbf{Game State: 2r2k2/1K1r1p2/5Q2/pp2pN2/P3P3/2P4P/1P3q2/8 w - - 3 46; Optimal Move: Kxc8} \\

\hrule
\scriptsize
\textbf{Experience Summary/Bootstrapping before RL}  

\nopagebreak\medskip{\color{grey!30!white}\hrule}\nopagebreak\medskip

The current game state is as follows: 2r2k21K1r1p25Q2pp2pN2P3P32P4P1P3q28

\textbf{Updated Understanding of the Game State}: 2r2k21K1r1p25Q2pp2pN2P3P32P4P1P3q28

**White to move.**

\textbf{Analysis of Simulation Trajectories} \textcolor{red}{(\textbf{re-stating the history})}:

1. **Trajectory 1: Kxc8**

   - **Initial Move**: Kxc8
   
   - **Opponent's Move**: f8g8
   
   - **Next State**: 2K3k1/3r1p2/5Q2/pp2pN2/P3P3/2P4P/1P3q2/8 w - - 1 47
   
   - **Subsequent Move**: Kb8
   
   - **Opponent's Move**: g8f8
   
   - **Next State**: 1K3k2/3r1p2/5Q2/pp2pN2/P3P3/2P4P/1P3q2/8 w - - 3 48

2. **Trajectory 2: Ka6**

   - **Initial Move**: Ka6
   
   - **Opponent's Move**: f8g8
   
   - **Next State**: 2r3k1/3r1p2/K4Q2/pp2pN2/P3P3/2P4P/1P3q2/8 w - - 5 47
   
   - **Subsequent Move**: Kxa5
   
   - **Opponent's Move**: g8f8
   
   - **Next State**: 2r2k2/3r1p2/5Q2/Kp2pN2/P3P3/2P4P/1P3q2/8 w - - 1 48

\textbf{Evaluation}:

1. **Kxc8 Trajectory**:
   - After Kxc8, the board simplifies, and the white king is now in a more exposed position. The black king is still a threat, and the board is less controlled by the white queen.
   - The subsequent move Kb8 does not significantly improve the position, and the black king continues to be a threat.

2. **Ka6 Trajectory**:
   - After Ka6, the white king is more protected, and the board is less open. The black king is still a threat, but the white queen has more control over the center and the board.
   - The subsequent move Kxa5 does not significantly improve the position, and the black king continues to be a threat. \textcolor{red}{(\textbf{Superficial evaluation})}

\textbf{Conclusion}:

While both moves have their merits, Ka6 provides better protection for the king and maintains more control over the board. Therefore, the updated understanding is that Ka6 is a better move to explore further.

**Next Move: Ka6**  \textcolor{red}{\textbf{(Surface-level evaluation, Over-indexed on “evaluation form” rather than chess substance)}}

\nopagebreak\smallskip
{\color{black!50!white}\tikz\draw[dash pattern=on 3pt off 2pt, line width=1.5pt] (0,0) -- (\linewidth,0);}
\nopagebreak\smallskip

\textbf{Experience Summary/Bootstrapping after RL} 

\nopagebreak\medskip{\color{grey!30!white}\hrule}\nopagebreak\medskip

The simulation history has .... 

Based on the simulation history, the position after the move Kxc8 would be a position with a queen on the board, while the position after the move Ka6 would be a position with a queen taken. Therefore, the position after the move Kxc8 is better for the white player.  \textcolor{ForestGreen}{\textbf{(Outcome-oriented, Evaluates the resulting material outcome)}}
\\

\hrule

\end{tcolorbox}

\begin{tcolorbox}[
  breakable,
  colback=grey!5!white,
  colframe=grey!50!white,
  listing options={
    basicstyle=\ttfamily\footnotesize,
    breaklines=true,
    breakatwhitespace=false,
    columns=fullflexible,
  },
    code={\refstepcounter{table}}, 
    title={Table \thetable: Reasoning trace example THREE of SKL-RL before and after end-to-end RL training.},
    label={table: chess sg-rl trained trace 3}
]

\footnotesize

\textbf{Game State}: r2r2k1/p4ppp/1p1R3b/n3p3/N7/2P2N2/P3K1PP/3R4 w - - 4 22
\medskip

\hrule
\medskip

\scriptsize
\textbf{State Understanding Before RL}

\nopagebreak\medskip{\color{grey!30!white}\hrule}\nopagebreak\medskip

The current position is as follows: r2r2k1/p4ppp/1p1R3b/n3p3/N7/2P2N2/P3K1PP/3R4 w - - 4 22

\textbf{Key Features}:

1. **White to Move**: White is in the process of checkmating Black.

2. **Material**: White has a rook, two knights, and a king. Black has a king and a rook.

3. **Positional Evaluation**:

   - White's rook is on the 6th rank, which is a strong position.
   
   - Black's rook is on the 7th rank, which is a weak position.
   
   - White's knights are well-placed to attack Black's king.
   
   - Black's king is in the center, with no immediate escape routes.

\textbf{Strategic Considerations}:

- **Checkmate Threat**: White has a strong checkmate threat with the rook and knights.

- **Rook Control**: The rook on the 6th rank is a key piece, controlling the 6th and 7th ranks.

- **Knight Maneuvering**: The knights can be used to deliver checkmate or put pressure on Black's king.

\textbf{Candidate Moves} \textcolor{red}{(\textbf{spamming})}:

1. **Rd8**: Checkmate with the rook.

2. **Rd7**: Checkmate with the rook.

3. **Rxh6**: Captures a piece and puts pressure on Black's king.

4. **Rg6**: Check with the rook, putting pressure on Black's king.

5. **Rf6**: Check with the rook, putting pressure on Black's king.

6. **Rc6**: Check with the rook, putting pressure on Black's king.

7. **Rxb6**: Captures a piece and puts pressure on Black's king.

8. **Nxb6**: Captures a piece and puts pressure on Black's king.

9. **Nc5**: Maneuvering the knight to a key square.

10. **Ng5**: Maneuvering the knight to a key square.

\textbf{Reasoning}:

- **Rd8**: This move is the most direct and immediate checkmate threat. If Black's king cannot escape, this will result in checkmate.

- **Rd7**: Similar to Rd8, but slightly less direct. It still puts pressure on Black's king and can lead to a quick checkmate.

- **Rxh6**: Captures a piece and puts pressure on Black's king. This move is less direct but still a strong threat.

- **Nxb6**: Captures a piece and puts pressure on Black's king. This move is also a strong threat but less direct than the rook moves.

\textbf{Conclusion}:

The best move is **Rd8**, as it directly leads to checkmate if Black's king cannot escape. This move is the most immediate and decisive.

**Next Move: Rd8    \textcolor{red}{\textbf{(Prefer to propose pure assertion, not conclusion, spamming candidate moves)}}

\nopagebreak\smallskip
{\color{black!50!white}\tikz\draw[dash pattern=on 3pt off 2pt, line width=1.5pt] (0,0) -- (\linewidth,0);}
\nopagebreak\smallskip

\textbf{State Understanding After RL} 

\nopagebreak\medskip{\color{grey!30!white}\hrule}\nopagebreak\medskip

This position is complex and dynamic, with White needing to take advantage of their material superiority and central control.

\textbf{Candidate moves}:

- 'Kd3': This is too passive and doesn't address the immediate threats.

- 'Rxd8': This sacrifices the rook but takes out a key black rook and opens up the board.

- 'Kf2': This centralizes the king and prepares to control key central squares.

After careful consideration, sacrificing the rook seems more aggressive and decisive.

'Rxd8' is the most dynamic move. It sacrifices the rook but takes out a key black rook and puts pressure on the black rook's central position.

Therefore, I'll go with:

Rxd8 \textcolor{ForestGreen}{\textbf{(More focused, Material awareness is actively considered)}, \textcolor{red}{\textbf{(but miss the checkmate advantage)}}}
\\

\hrule


\end{tcolorbox}

\subsection{Reasoning trace for SKL-RL with self-distillation loss }\label{appendix:skl-rl distill trace}

\begin{tcolorbox}[
  breakable,
  colback=grey!5!white,
  colframe=grey!50!white,
  title=Reasoning trace of SKL-RL after adding self-distillation objective function $\mathcal{L}_{\text{Distill}}$.
]

\scriptsize

\textbf{Game State}: \\
r1b1k2r/5ppp/2p5/p1q1p1B1/Pp2P1n1/1P3NN1/1P3PPP/3Q1RK1 w kq - 0 17

\medskip

\hrule
\medskip

\textbf{[State Understanding]}\\

Based on the simulation history, our understanding of the game state is as follows: \textcolor{blue}{(\textbf{imprint from forward KL self-distillation data})} \\

1. Action \textbf{Qd8} appears to be the best action among the three. It resulted in a perfect game termination with a reward of 1.0. This action clearly outperformed the other two actions in terms of long-term value. The simulation history for \textbf{Qd8} is overwhelmingly positive, and it is the only action that led to a game termination with a perfect reward. Therefore, we can confidently conclude that Qd8 is the best action to take in this game state. We should not consider this action in future game play as it has demonstrated superior performance and potential for future success. \textcolor{ForestGreen}{\textbf{(reasoning with the distilled experience)}} \\

2. Action \textbf{Qd7} did not perform well in the simulation history. Although it did not result in a game termination, it did not yield any positive reward. The simulation history for \textbf{Qd7} did not provide any evidence of future value. In fact, the game terminated with a reward of 0.0, indicating that the action did not lead to a successful outcome. The simulation history for \textbf{Qd7} did not provide any evidence of future value, and it did not perform as well as \textbf{Qd8} in terms of long-term value. Therefore, we can conclude that \textbf{Qd7} is not a good action to take in this game state.\\

3. Action \textbf{Qd6} ...[Same inference as 2]. However, we should not completely rule out the possibility that Qd6 could perform better in future game play. We should continue to consider Qd6 as a potential action in future game play, but we should not give it the same level of consideration as Qd8. We should be cautious in our use of Qd6 and only consider it as a potential action when Qd8 is not available. We should also be aware that Qd6 did not perform as well as Qd7 in terms of long-term value. Therefore, we should give Qd6 the lowest level of consideration among the three actions. \textcolor{ForestGreen}{\textbf{(careful and reflective about its own decisions)}} \\

\textcolor{ForestGreen}{\textbf{[In Brief]: reasoning with previous knowledge, but not over-confident.}} \\

\nopagebreak\smallskip
{\color{black!50!white}\tikz\draw[dash pattern=on 3pt off 2pt, line width=1.5pt] (0,0) -- (\linewidth,0);}
\nopagebreak\smallskip

\textbf{[Bootstrapping / Self-Evaluation]} \\

Based on the simulation history, my updated understanding of the game state is as follows:\\

\#\#\# Detailed Analysis \textcolor{ForestGreen}{\textbf{(Recognize multiple subsequent steps in simulation)}}:

1. \textbf{Move Qd8}: 

- Result: Terminated the game with a reward of 1.0. 

- \textbf{Conclusion}: Qd8 is the best move and the only move that leads to a win. It is the optimal move in this state.

2. \textbf{Move Qd7}:

   - Opponent's move e8f8 led to a state where the game did not terminate.
   
   - Subsequent moves Qxg4, Qd2, and Rd1 all led to states where the game did not terminate. \textcolor{ForestGreen}{(\textbf{More explicit trajectory analysis})}
   
   - \textbf{Conclusion}: Qd7 is not a winning move, and it does not guarantee a win. It is a suboptimal move.

3. \textbf{Move Qd6}:

   - Opponent's move h8g8 led to a state where the game did not terminate.
   
   - Subsequent moves Qd2 and Ra1 both led to states where the game did not terminate.
   
   - \textbf{Conclusion}: Qd6 is not a winning move, and it does not guarantee a win. It is a suboptimal move.

4. \textbf{Move Qd5}:

   - Opponent's move h8g8 led to a state where the game did not terminate.
   
   - Subsequent moves Qd6 led to states where the game did not terminate.
   
   - \textbf{Conclusion}: Qd5 is not a winning move, and it does not guarantee a win. It is a suboptimal move.\\

\#\#\# Summary: \textcolor{ForestGreen}{\textbf{(Well-rounded consideration)}}

- \textbf{Qd8} is the only move that guarantees a win and terminates the game with a reward of 1.0.

- \textbf{Qd7, Qd6, and Qd5} do not guarantee a win and do not terminate the game. They all lead to states where the game continues and does not terminate.\\






Therefore, the updated understanding is:

\(\boxed{Qd8}\) is the best move to play in this state. Qd7, Qd6, and Qd5 are suboptimal moves that do not guarantee a win. Qd8 is the only move that solves the game.  \\

\end{tcolorbox}

\end{document}